%% file: 00_Paper_Top_IEEE.tex
\documentclass[journal]{IEEEtran}
\usepackage{amsmath,amssymb}
\usepackage{relsize}
\usepackage{xcolor}
\usepackage{MnSymbol}
\usepackage[mathscr]{euscript}

\usepackage{amsthm}

\newtheorem{definition}{Definition}
\newtheorem{theorem}{Theorem}

\usepackage{listings}

\usepackage{mathtools}

\usepackage{makecell}
\setcellgapes{4pt}

\usepackage{algorithmic}
\usepackage[ruled,noline,linesnumbered]{algorithm2e}

\usepackage[justification=centering]{caption}

\usepackage{pifont}

\usepackage{cite}

\usepackage{mdwlist,url}
\usepackage{url}

\ifCLASSINFOpdf
\usepackage[pdftex]{graphicx}

\graphicspath{{Img/}}

\else
\fi

\usepackage{amsmath}  
\usepackage{float}
\usepackage{subfig}
\input{commenting}

\usepackage{comment}

\usepackage{longtable}
\newcounter{magicrownumbers}

\DeclareMathOperator*{\argmax}{arg\,max}

\begin{document}
	\title{Providing Error Detection for Deep Learning Image Classifiers Using Self-Explainability}

\newcommand{\mmk}[1]{{\footnotesize\color{purple}[MMK: #1]}}
	
	\author{
		\IEEEauthorblockN{MohammadMahdi Karimi, Azin Heidarshenas, William W. Edmonson\\}
	}

	\IEEEoverridecommandlockouts
	\IEEEpubid{\makebox[\columnwidth]{
			\copyright2022
			\hfill} \hspace{\columnsep}\makebox[\columnwidth]{ }}

	\markboth{
}%
	{Shell \MakeLowercase{\textit{et al.}}: Bare Demo of IEEEtran.cls for IEEE Journals}
	
	\maketitle
	\begin{abstract}
    \input{001_abs}

	\end{abstract}
	
	
	\IEEEpeerreviewmaketitle

\input{01_intro}
\input{02_related}

\input{03_background}

\input{04_proposed_se_dl}

\input{05_concept_selection}
\input{06_eval_methodology}
\input{07_results}

\input{08_conclusion}

	\bibliographystyle{IEEEtran}
	\bibliography{00_Paper_Top_IEEE.bib} 
\vspace{-.5in}	
\begin{IEEEbiographynophoto}{MohammadMahdi Karimi} received his Ph.D. in Electrical and Computer Engineering from North Carolina A\&T State University. He is currently an Automotive Safety Engineer at NVIDIA Corporation, CA, USA. 
His research interests include safety of deep learning systems, and graph neural networks. 
\end{IEEEbiographynophoto}	
\vspace{-.95in}
\begin{IEEEbiographynophoto}{Azin Heidarshenas} received her Ph.D. in Electrical and Computer Engineering from University of Illinois Urbana-Champaign. She is currently a Machine Learning Engineer at Apple Inc., CA, USA. 
Her research interests include hardware acceleration of graph applications, and safety of deep learning systems. 
\end{IEEEbiographynophoto}	
\vspace{-.95in}
\begin{IEEEbiographynophoto}{William W. Edmonson} 
is currently the founder of Sadaina LLC (Space Technology Company) and MaxIQ (Space STEM Education). He is a retired Professor from the Electrical and Computer Engineering Department at NC A\&T State University (NCAT). He received his Ph.D. in Electrical and Computer Engineering from NC State University in 1990. From 2010-19, he was the National Institute of Aerospace S.P. Langley Professor with the Department of Electrical and Computer Engineering, NCAT, and the Director of the Small Satellite Systems Research Center, whose research focused on advancing the capabilities, functionality, and scope of missions for small satellites, particularly in the areas of inter-satellite communication and verifiable model-based systems engineering.
Dr.~Edmonson serves on the AIAA Small Satellite Technical Committee and is a Senior Member of the Institute of Electrical and Electronics Engineers (IEEE).

\end{IEEEbiographynophoto}
\end{document}

%% file: commenting.tex
\definecolor{WowColor}{rgb}{.75,0,.75}
\definecolor{SubtleColor}{rgb}{0,0,.50}

\newcounter{margincounter}

%% file: 001_abs.tex
This paper proposes a self-explainable Deep Learning (SE-DL) system for an image classification problem that performs self-error detection. The self-error detection is key to improving the DL system's safe operation, especially in safety-critical applications such as automotive systems. A SE-DL system outputs both the class prediction and an explanation for that prediction, which provides insight into how the system makes its predictions for humans. Additionally, we leverage the explanation of the proposed SE-DL system to detect potential class prediction errors of the system. The proposed SE-DL system uses a set of concepts to generate the explanation. The concepts are human-understandable lower-level image features in each input image relevant to the higher-level class of that image. We present a concept selection methodology for scoring all concepts and selecting a subset of them based on their contribution to the error detection performance of the proposed SE-DL system. Finally, we present different error detection schemes using the proposed SE-DL system to compare them against an error detection scheme without any SE-DL system.


%% file: 01_intro.tex
\section{Introduction}

Deep Learning~(DL) systems are increasingly being used in many safety-critical automotive applications such as image classification systems. Although they provide fast predictions at high classification accuracy in most cases; however, their pervasive deployment would not be possible without ensuring their correct functionality. 

Previous research has shown that lack of diversity in the training dataset or selection of a very large DL network architecture with a substantial number of weights can lead to poor training~\cite{tabernik2019deep, shustanov2017cnn}. As a result, the system may fail to predict unseen data and lose its robustness against perturbed inputs. Such inadequate performance can lead to unprecedented system malfunction and failures due to reduced classification accuracy when the model is deployed in the real-world environment. Examples of such failures include uncertainty in the DL system's prediction and their brittleness to adversarial examples~\cite{choi20217}. Consecutively, potentially hazardous driving incidents can occur that harm humans or cause damage to property or the environment. Therefore, in safety-critical applications, it is crucial to detect potential malfunctions in the system to prevent them from causing hazards.

However, since DL systems are black-box~\cite{rudin2019stop, adadi2018peeking, varshney2016engineering}, it is challenging to detect such potential malfunctions in the system and determine if the DL system's prediction is trustworthy~\cite{ribeiro2016should, elton2020self, balayan2020teaching}. The \textit{explanation} techniques provide human-understandable insight on how the DL system makes its predictions~\cite{ghorbani2019towards, ming2019interpretable, ribeiro2016should}.

This paper proposes a method to develop a self-explainable Deep Learning (SE-DL) system for a single-class image classification problem that generates explanation along with its output class prediction. Additionally, we leverage the explanation of the proposed SE-DL system to detect potential class prediction errors of the system. The proposed SE-DL system uses a set of concepts to generate the explanation. The concepts are human-understandable lower-level image features in each input image relevant to the higher-level class of that image. We also propose a concept selection methodology to score all concepts and select a subset of them based on their contribution to the error detection performance of the proposed SE-DL system. 
Finally, we present different error detection schemes using the proposed SE-DL system to compare them against an error detection scheme without any SE-DL system of~\cite{pang2019improving, sen2020empir, wei2020cross}.

%% file: 02_related.tex
\section{Related Work}

Several research works have proposed explanation techniques for Deep Learning~(DL) systems, particularly image classification systems~\cite{li2018deep, kindermans2019reliability, elton2020self}. 
An explanation technique reveals how the DL system makes predictions, e.g., by providing the most important input data features that contribute to its prediction~\cite{ribeiro2016should, ghorbani2019towards, wang2020score}. 
However, since DL systems are black-box, it is unknown when the system makes erroneous predictions. Since the explanation provides insight into how a DL system predicts, they are essential to detect erroneous predictions.

Unfortunately, current explanation techniques cannot be used to perform error detection in DL systems. Generally, there are two types of explanation techniques for DL systems: post-hoc explanation DL systems and self-explainable DL systems. Post-hoc techniques use an external system to generate the explanation in the form of saliency maps, which are the visual representation of the most important input features that explain the response of a DL system to humans~\cite{xie2020explainable, kindermans2019reliability}. Although post-hoc techniques can be useful in explaining the prediction of DL systems, it is unclear how saliency maps can be leveraged to detect when the DL system makes an erroneous prediction. Additionally, a post-hoc technique still relies on a different system to generate the saliency maps, which itself may malfunction.

In contrast, self-explainable DL systems~\cite{li2018deep, alvarez2018towards} do not rely on an external system to provide an explanation, and the system itself generates explanation along with prediction.  Current self-explainable systems rely on the generated explanation to make the prediction. Therefore, if the explanation is wrong, the prediction might become erroneous. As a result, it is impractical to detect when the system makes erroneous predictions using such self-explainable techniques.

Additionally, ensemble techniques~\cite{pang2019improving, sen2020empir, wei2020cross} train multiple independently developed DL systems and use a majority vote mechanism to generate the final prediction. Ensemble techniques can detect prediction errors by comparing the predictions of individual DL systems. Using ensemble techniques, one can detect a higher number of prediction errors by incorporating a higher number of DL systems and comparing their predictions~\cite{sen2020empir, wei2020cross, tramer2018ensemble, pang2019improving}. However, the downside of ensemble techniques is that training a large number of DL systems induces high computational complexity. As a result, there is a trade-off between computational complexity and the number of detected errors in ensemble techniques. In fact, to build an ensemble classification system, most research works incorporate between three~\cite{tramer2018ensemble, sen2020empir} to five~\cite{wei2020cross, pang2019improving} DL classifiers to increase the number of detected errors while avoiding high computational complexity. 

In this paper, we propose a self-explainable DL  single-class image classification system that is capable of detecting \textit{class prediction errors} without relying on additional classifiers in an ensemble setting. 

%% file: 03_background.tex

\section{Basic Definitions}\label{sec:basic_def}~
Before we describe the details of the proposed self-explainable deep learning~(SE-DL)  system design, we first provide some basic definitions that we will use throughout this paper. 
In an image classification problem, the input image $x\in \mathbb{R}^d$ is a vector of size $d$ (number of pixels across three RGB color channels) with real values between 0 and 1. The DL system categorizes the input $x$ into a specific \textit{class}, which we define as follows:

\begin{definition}\label{def:class}
\textbf{Set of Classes:} $\mathbb{C}:\{C_1,\dots,C_N\}$ is a set of classes identified in the image classification problem's dataset. In a single-class image classification problem, the dataset assigns one class to each image which is based on the content of that image.
\end{definition}

\begin{figure}
	\begin{center}
	\begin{minipage}{\columnwidth}
        \centering

        \subfloat[``Prohibited for all vehicles''
        \label{fig:cls_cncpt_class}]{\includegraphics[page={1}, width=0.35\columnwidth]{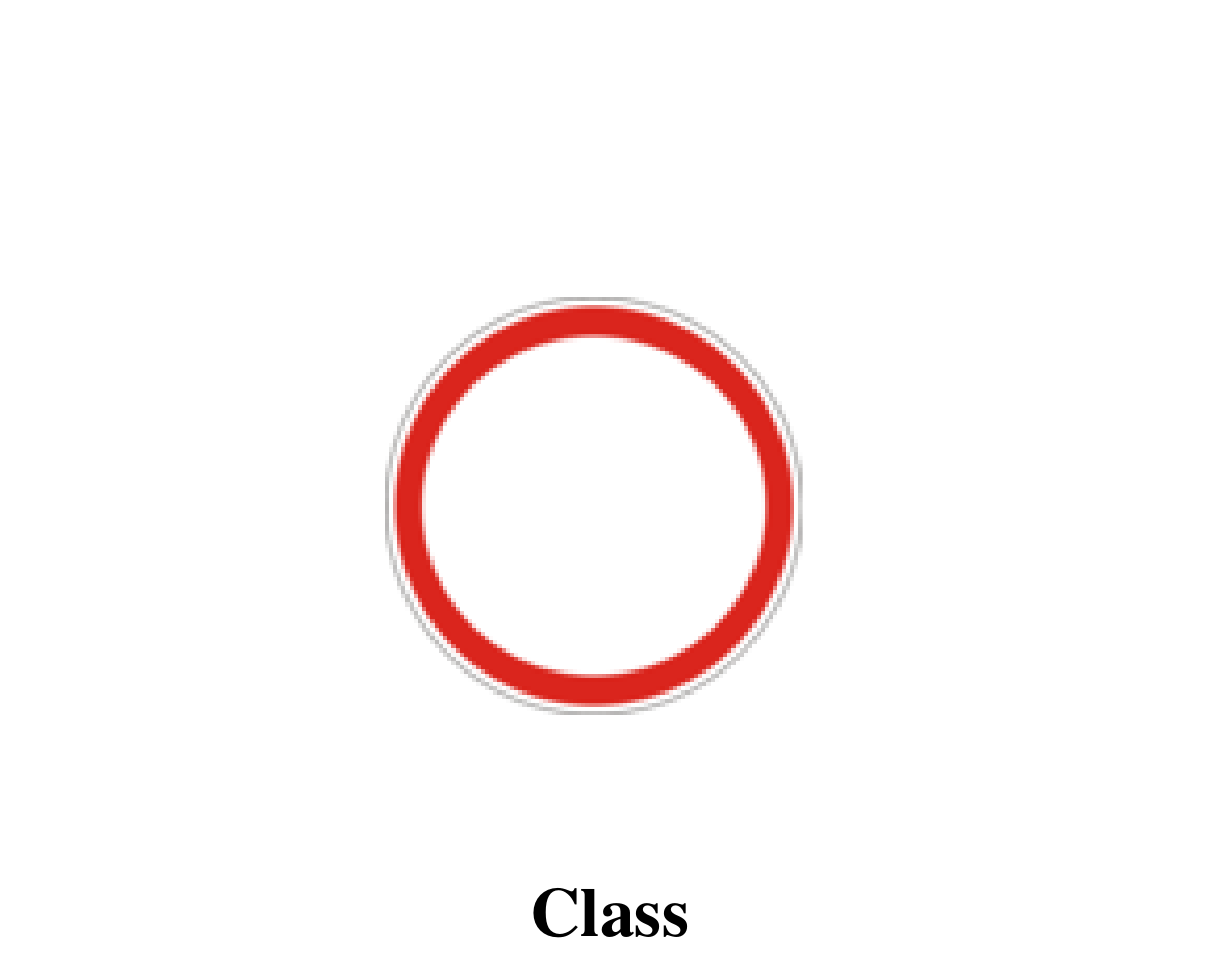}}
        \subfloat[$E(\text{``Prohibited for all vehicles''})$
        \label{fig:cls_cncpt_concept}]{\includegraphics[page={2}, width=0.35\columnwidth]{Img/cls_concept.pdf}}
        \caption{Example of class, concepts and explanation in GTSRB~\cite{Houben-IJCNN-2013} image classification problem.}
        \label{fig:cls_cncpt}
	\end{minipage}
	\end{center}
\end{figure}

For a given dataset, we visually identify specific colors or shapes, which are common lower-level image features that inform higher-level classes in a classification problem, defined as follows:

\begin{definition}\label{def:concept}
\textbf{Concept:} 
In an image classification problem, \textit{concepts} are human-understandable lower-level image features that are relevant to the higher-level class, which the image belongs to. We define each such image feature as concept $a_i \in \mathbb{A}:\{a_1,\dots , a_M\}$.
\end{definition}

For example, the ``Prohibited for all vehicles'' traffic sign class from German Traffic Sign Recognition Benchmark (GTSRB)~\cite{Houben-IJCNN-2013} shown in Figure~\ref{fig:cls_cncpt} contains concepts ``Shape circle'' and ``Color red''    (Figure~\ref{fig:cls_cncpt_concept}). We use the next definition to specify the relationship between a class and a concept:

\begin{definition}\label{def:concept_explains}
\textbf{Class-Concept Relationship:} A class $C_j\in\mathbb{C}:\{C_1,\dots,C_N\}$ is associated with a concept $a_i\in\mathbb{A}$ or similarly, the concept $a_i$ explains class $C_j$, if the concept exists in the image instances of that class in a given dataset.
\end{definition}

In the example shown in Figure~\ref{fig:cls_cncpt}, the ``Prohibited for all vehicles'' class is associated with concepts ``Shape circle'' and ``Color red''. Similarly, the concepts ``Shape circle'' and ``Color red'' explain the class ``Prohibited for all vehicles''. For each class $C_j$, we define the explanation function as follows:

\begin{definition}\label{def:explanation}
\textbf{Explanation $E(C_j)$:} An explanation $E(C_j):\mathbb{C}\rightarrow \mathcal{P}(\mathbb{A})$ is a mapping between classes and the power set of concepts $\mathcal{P}(\mathbb{A})$, which includes all the subsets of $\mathbb{A}$. The explanation function is derived from the dataset and determines the set of concepts that explain each class.  
\end{definition}
In the example shown in Figure~\ref{fig:cls_cncpt}, we have $E\big($``Prohibited for all vehicles''$\big)=\big\{$ ``Shape circle'', ``Color red'' $\big\}$.

%

%% file: 04_proposed_se_dl.tex
\section{Proposed Method}\label{sec:proposed_method}
\begin{figure}[t]
    \centering
        \centering
        \vspace{-1.5cm}\includegraphics[width=0.9\linewidth]{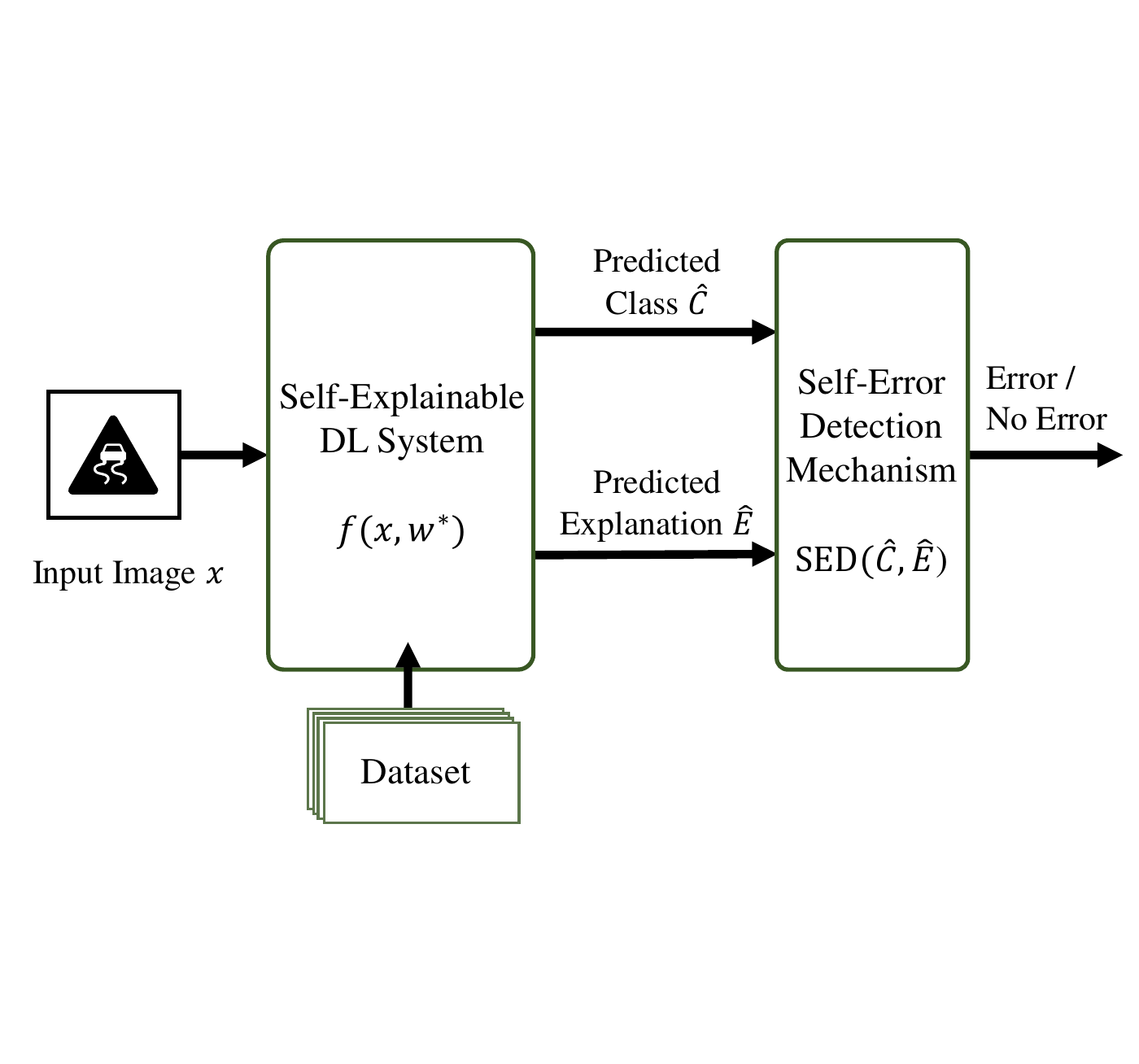}
        \vspace{-1.2cm}\caption{Overview of the proposed SE-DL system and self-error detection mechanism.}
        \label{fig:main}
        
\end{figure}

\begin{figure}[t]
        \centering
        \includegraphics[width=0.9\linewidth]{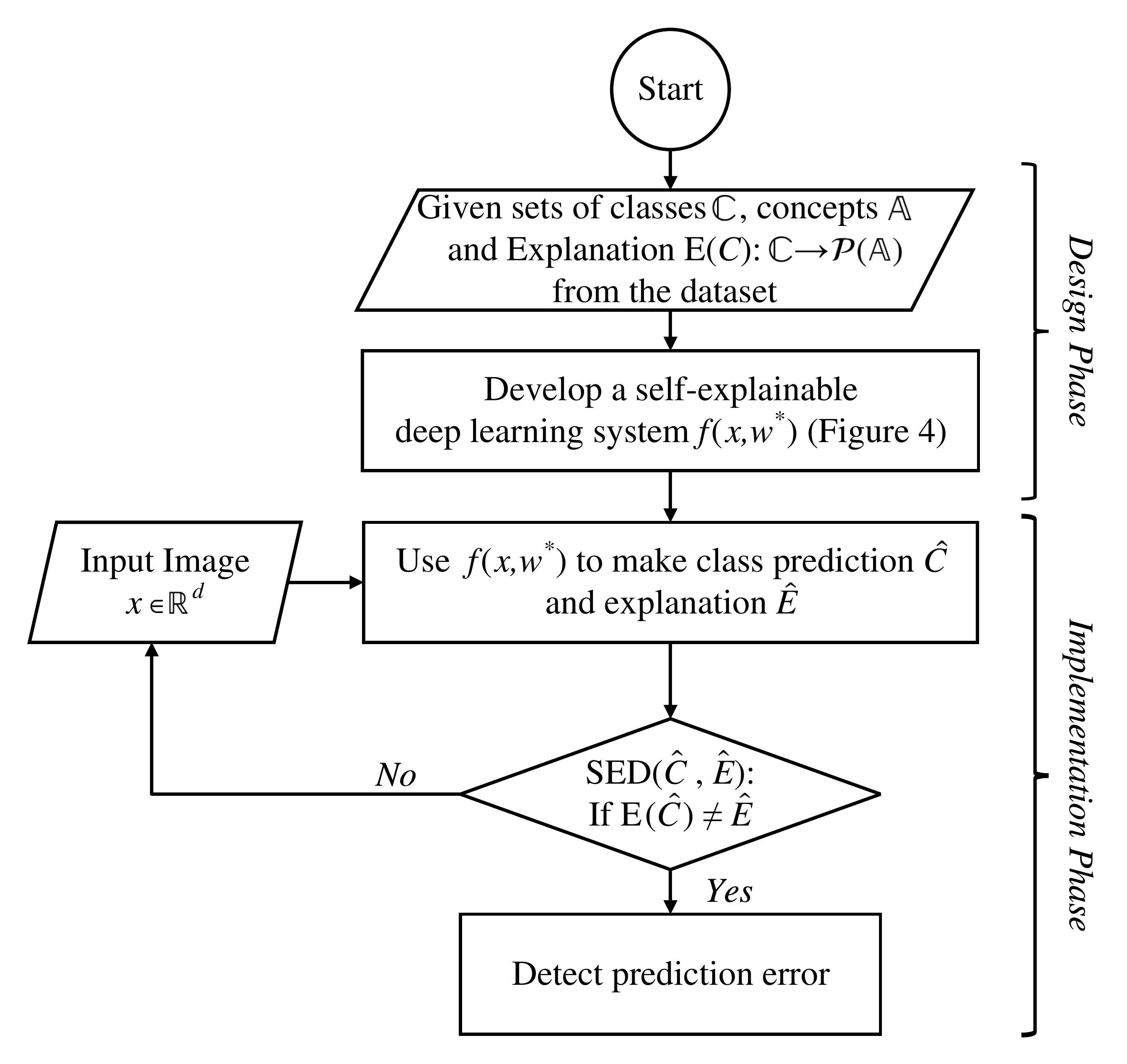}
        \caption{Workflow of the proposed SE-DL system and self-error detection mechanism.}
        \label{fig:overall_flow}
\end{figure}

~
Figures~\ref{fig:main} and~\ref{fig:overall_flow} illustrate the overview and the workflow, respectively, of the proposed self-explainable deep learning~(SE-DL) image classifier system and the proposed self-error detection mechanism. During the design phase, given sets of classes $\mathbb{C}$, concepts $\mathbb{A}$, and explanation function $E(C):\mathbb{C}\rightarrow\mathcal{P}(\mathbb{A})$ from the dataset, we develop the proposed SE-DL system $f(x, w^*)$, which we will discuss in Subsection~\ref{sec:prop_self_exp_sys}. The SE-DL system takes as input an image $x$ and outputs both the predicted class $\hat{C}$ and the predicted explanation $\hat{E}$. 
Furthermore, we leverage the proposed SE-DL system to design a self-error detection mechanism $\text{SED}(\hat{C}, \hat{E})$, which we will discuss in Subsection~\ref{sec:prop_self_err_det_mech}. The self-error detection mechanism detects takes as input both of the outputs from the SE-DL system and generates an ``error'' or a ``no error'' signal given image $x$ during the implementation phase.


\subsection{Proposed Self-Explainable Deep Learning System}\label{sec:prop_self_exp_sys}~
We define our problem as designing a SE-DL system for a single-class image classification problem. We reformulate this problem into \textit{a new multi-class classification problem} that predicts both the class $\hat{C}$ and the explanation $\hat{E}$ of an input image $x$. We define the proposed SE-DL system for a single-class image classification problem as follows:
\begin{definition}\label{def:self_exp}
\textbf{Self-Explainable DL (SE-DL) System}  $f(x,w^*):\mathbb{R}^d\rightarrow \mathbb{C}\times \mathcal{P}(\mathbb{A^*})$ takes as input an image $x$ and generates as output $<\hat{C},\hat{E}>$. The $\hat{C}\in \mathbb{C}$ and $\hat{E}\in \mathcal{P}(\mathbb{A}^*)$ are the predicted class and the predicted explanation, respectively. The $\mathbb{A}^*\subseteq \mathbb{A}$ is a minimal set of selected concepts used to develop the SE-DL system (which we elaborate in Section~\ref{sec:concept-selection}), and $w^*$ includes the weights of the trained model. 
\end{definition}

From Definition~\ref{def:self_exp}, we construct the proposed SE-DL system architecture $f(x,w)=<\hat{C},\hat{E}>$  to have $N+M^*$ predictions, where $N$ is the number of classes and $M^* = |\mathbb{A}^*|$ is the number of selected concepts used for the explanation. Consequently, we write the loss function of the multi-class image classification problem with $N+M^*$ predictions as Equation~\ref{eq:main-ova}:
\begin{equation}\label{eq:main-ova}
\smaller
    \begin{aligned}
        &loss\big(f(x,w),y\big) =\\&-\frac{1}{N + M^*} \mathlarger{\mathlarger{\sum}}_{i=0}^{N + M^*}y[i]\times \text{log}\Big(\frac{\text{exp}\big(f(x,w)[i]\big)}{1+\text{exp}\big(f(x,w)[i]\big)}\Big) + \\
	&(1-y[i])\times\text{log}\Big(\frac{1}{1+\text{exp}\big(f(x,w)[i]\big)}\Big)
    \end{aligned}
\end{equation}
where $y\in \{0,1\}^{N+M^*}$ is a vector of size $N+M^*$ that contains ground truth class $C_x\in \mathbb{C}$ of input image $x$ and the ground truth explanation (given from the dataset) of the input image $E(C_x)\in \mathcal{P}(\mathbb{A}^*)$, i.e., $y=<C_x,E(C_x)>$. 
Finally, the proposed SE-DL system is trained to obtain $f(x,w^*)$ using a training dataset, and the weights are updated using the Stochastic Gradient Descent technique~\cite{goodfellow2016deep}.

\subsection{Proposed Self-Error Detection Mechanism}\label{sec:prop_self_err_det_mech}~
We leverage the proposed SE-DL system to design a self-error detection mechanism to detect class prediction errors. A class prediction error occurs when for a given input image $x$, the predicted class output of the DL system does not match the ground truth class of the input image, i.e., $\hat{C}\neq C_x$. 
According to Figure~\ref{fig:main}, the self-error detection mechanism takes as input both the predicted class $\hat{C}\in\mathbb{C}$ and the predicted explanation $\hat{E}\in \mathbb{A}^*$ and generates an error signal if the prediction is erroneous. The self-error detection mechanism generates an error signal based on Definition~\ref{def:sed}: 

\begin{definition}\label{def:sed}
\textbf{Self-Error Detection Mechanism} $\emph{SED}(\hat{C},\hat{E})$\textbf{:} Given a trained SE-DL system $f(x,w^*)=<\hat{C},\hat{E}>$ (Definition~\ref{def:self_exp}), we detect class prediction errors using self-error detection function $\emph{SED}(\hat{C},\hat{E})$ as follows:
\begin{equation}\label{eq:self_err_det_mech}
    \emph{SED}(\hat{C},\hat{E}) = 
    \begin{cases}
    {True}\:\: (\emph{Error}) & if  \hat{E} \neq E(\hat{C})\\
    {False}\:\: (\emph{No Error}) & if \hat{E} = E(\hat{C})
    \end{cases}
\end{equation}
where $E(\cdot): \mathbb{C}\rightarrow\mathcal{P}(\mathbb{A}^*)$ is the explanation function given from the dataset (Definition~\ref{def:explanation}).
\end{definition}
Specifically, the self-error detection mechanism detects a class prediction error if the predicted explanation $\hat{E}$ does not match the ground truth explanation of the predicted class $E(\hat{C})$. 
If a concept is present or absent in the ground truth explanation of the predicted class $E(\hat{C})$, it needs to be present or absent in the predicted explanation $\hat{E}$; otherwise, a class prediction error is detected using the proposed self-error detection mechanism. 
In the example of Figure~\ref{fig:cls_cncpt}, if the predicted class is $\hat{C}=$``Prohibited for all vehicles'', where $E(\hat{C})=\big\{$``Shape circle'', ``Color red''$\big\}$, the self-error detection mechanism detects an error if the concepts ``Shape circle'' or ``Color red'' are absent in $\hat{E}$ or if any other concept is present in $\hat{E}$.


The self-error detection mechanism of Definition~\ref{def:sed} only detects class prediction errors for which the ground truth explanation of the predicted class does not match the predicted explanation, i.e., $E(\hat{C})\neq \hat{E}$. We evaluate the performance of an error detection mechanism using images in a dataset $\forall x \in \mathbb{D}:\{(x, C_x)\}$, where $C_x\in \mathbb{C}$ is the ground truth class of the image $x$ specified in the dataset. 
To obtain all class prediction errors over the dataset, for each image $x$ we compare the ground truth class $C_x$ with the predicted class output $\hat{C}$ of the image classifier DL system $f(x,w^*)$. Specifically, given a dataset $\mathbb{D}$ of images, the set of all class prediction errors are computed in Equation~\ref{eq:all_err}:
\begin{equation}\label{eq:all_err}
    Err_{Total} = \big\{ \forall x\in \mathbb{D}, (C_x, \hat{C}) \:\:\:| \:\:\: C_x\neq \hat{C} \big \}
\end{equation}
By using an error detection mechanism we detect a subset of  all class prediction errors, $Err_{Total}$, in Equation~\ref{eq:all_err}. In the case of the self-error detection mechanism $\text{SED}(\hat{C},\hat{E})$ (Definition~\ref{def:sed}) we obtain the set of detected class prediction errors, $Err_{Detected}$, as Equation~\ref{eq:all_det_err}:
\begin{equation}\label{eq:all_det_err}
    Err_{Detected} = \big\{\forall x\in \mathbb{D}, (C_x, \hat{C}) \:\:\:| \:\:\: C_x\neq \hat{C}, \:\:\text{SED}(\hat{C},\hat{E}) = {True}\big \}
\end{equation}

We define the performance of an error detection mechanism $P_{ed}$ given a dataset as follows:
\begin{definition}\label{def:perf_err_det}\textbf{Performance of an Error Detection Mechanism $P_{ed}$} is defined as the ratio of the number of detected class prediction errors over the total number of class prediction errors given a dataset, as shown in Equation~\ref{eq:perf_err_det}:
\begin{equation}\label{eq:perf_err_det}
    P_{ed} := \frac{|Err_{Detected}|}{|Err_{Total}|}
\end{equation}
where $|\cdot|$ denotes the cardinality of a set.
\end{definition}
A higher $P_{ed}$ indicates that the error detection mechanism is more effective and larger number of error are detected using the error detection mechanism. In Section~\ref{sec:err_detection_scheme}, we present different error detection schemes based on the proposed SE-DL system and previous work (ensemble techniques~\cite{tramer2018ensemble, sen2020empir, wei2020cross}) and compare their error detection performance using Equation~\ref{eq:perf_err_det}.

\begin{figure}
	\begin{center}
		\includegraphics[width=0.65\columnwidth]{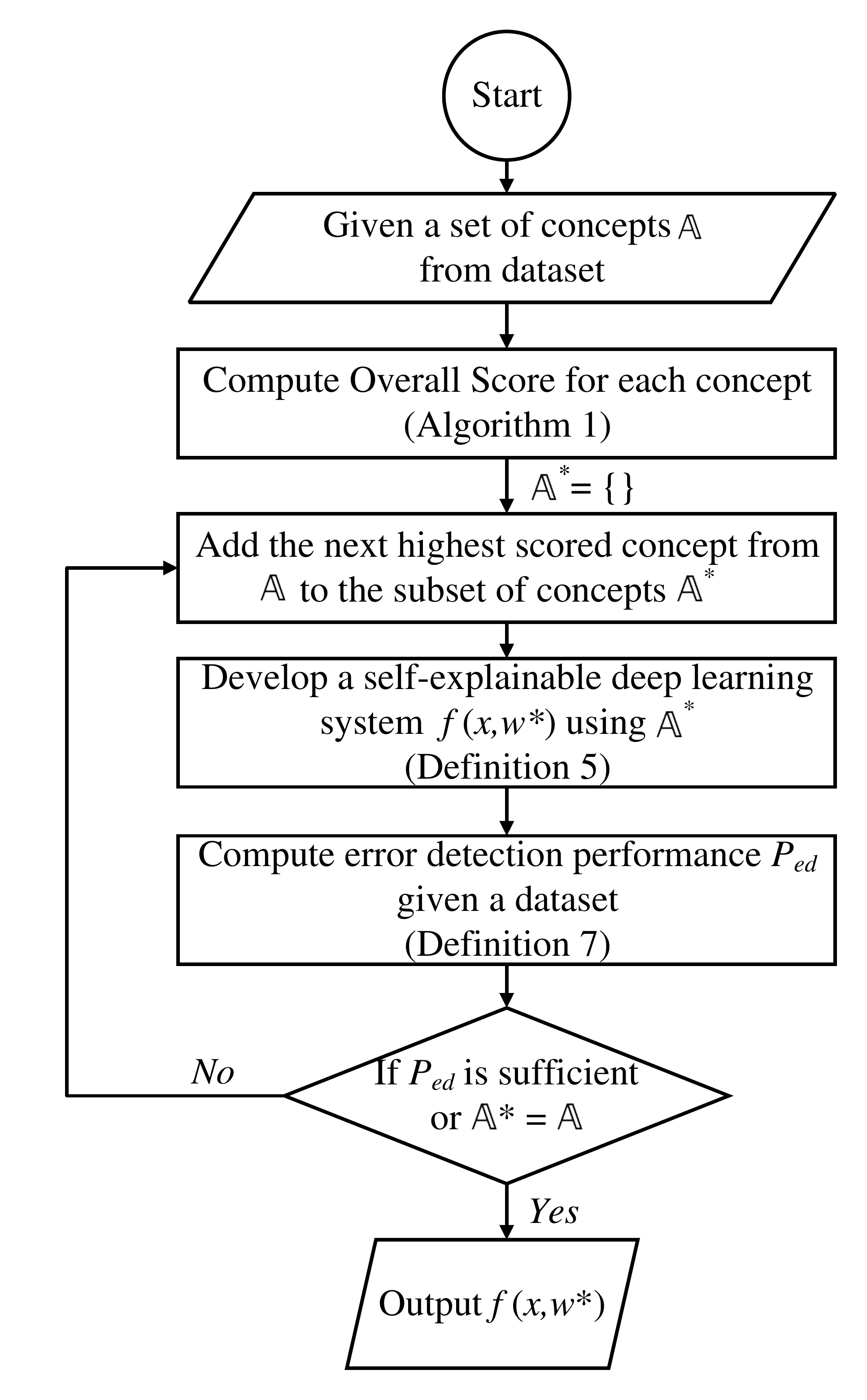}
		\caption{Design workflow of the proposed self-explainable DL system.}
		\label{fig:concept_overall_flow}
	\end{center}
\end{figure}

\subsection{Design Workflow}\label{sec:providing_sed}~
Figure~\ref{fig:concept_overall_flow} illustrates the design workflow of the proposed SE-DL system. Given a set of concepts $\mathbb{A}=\{a_1,\dots,a_M\}$ from the dataset, we first compute an overall score for each concept, which we will elaborate on in Algorithm~\ref{alg:concept_selection} in Subsection~\ref{sec:overall_score_concepts}. 

After computing the overall scores of all concepts, we select a subset of highest-scored concepts $\mathbb{A}^*\subseteq \mathbb{A}$. To compute the $\mathbb{A}^*$, we initialize $\mathbb{A}^*=\{\}$ and iteratively add concepts to it. In each iteration, we follow three steps. First, we add the next highest-scored concept to $\mathbb{A}^*$. Second, we develop the proposed SE-DL system $f(x,w^*)$ given the current $\mathbb{A}^*$, and use the SE-DL system to build a self-error detection mechanism (Definition~\ref{def:sed}). Third, we compute the error detection performance of the self-error detection mechanism as defined in Equation~\ref{eq:self_err_det_mech}. The workflow terminates and outputs the developed SE-DL system $f(x,w^*)$,
if the performance of the error detection mechanism is satisfactory or if all the concepts have been included, i.e., $\mathbb{A}^*$ has become equal to $\mathbb{A}$.

%% file: 05_concept_selection.tex
\section{Concept Selection}\label{sec:concept-selection}~
Thus far, we have presented the design of the proposed self-explainable deep learning~(SE-DL) system and the self-error detection mechanism. A key prerequisite for designing the proposed SE-DL system is selecting concepts that we can leverage to detect a large number of class prediction errors. In this section, we present a methodology for scoring concepts $\in \mathbb{A}$ to select a minimal subset of concepts $\mathbb{A}^*\subseteq \mathbb{A}$ based on their contribution to the error detection performance of the proposed SE-DL system.

\begin{figure}
	\begin{center}
	\begin{minipage}{0.35\textwidth}
	    \subfloat[Classes\label{fig:four_cls_class_example}]{\includegraphics[page={1}, width=\linewidth ]{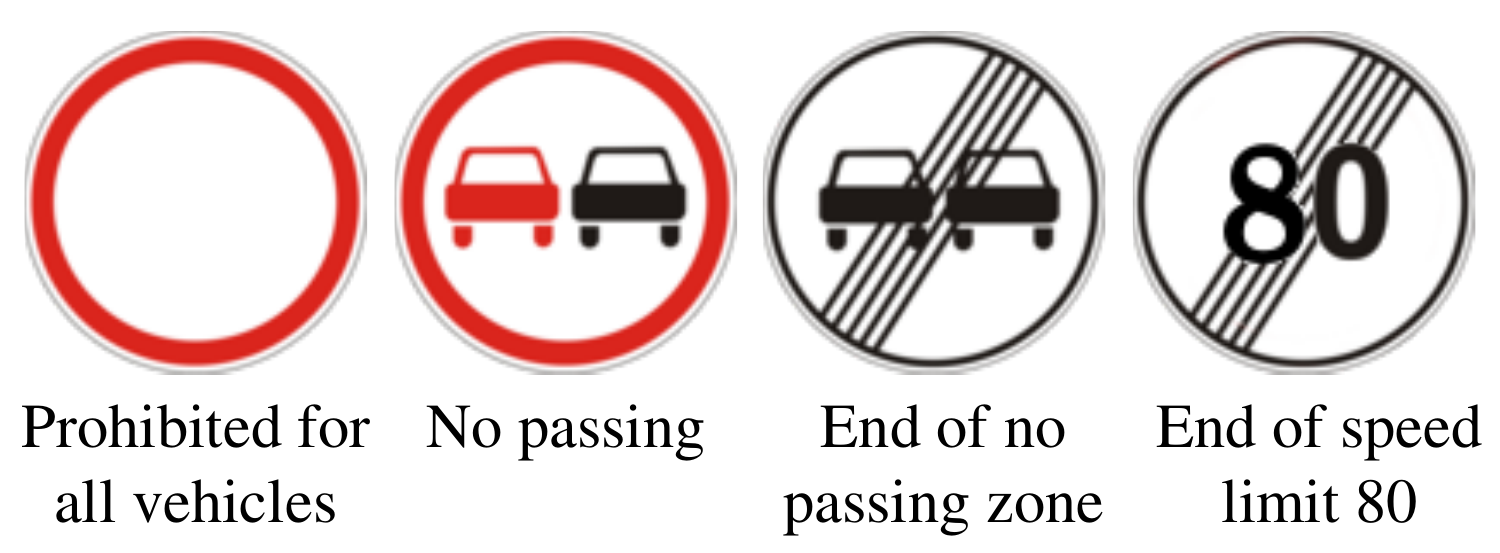}}
	    \\
	    \subfloat[Concepts\label{fig:four_cls_concept_example}]{\includegraphics[page={2}, width=\linewidth ]{Img/four_cls_cncpt_example.pdf}}	
    \end{minipage}
	   \hspace{1cm}
   \begin{minipage}{0.4\textwidth}
        \centering
        \subfloat[Associated classes with each concept
        \label{fig:four_cls_g_example}]{\includegraphics[page={1}, width=0.85\linewidth]{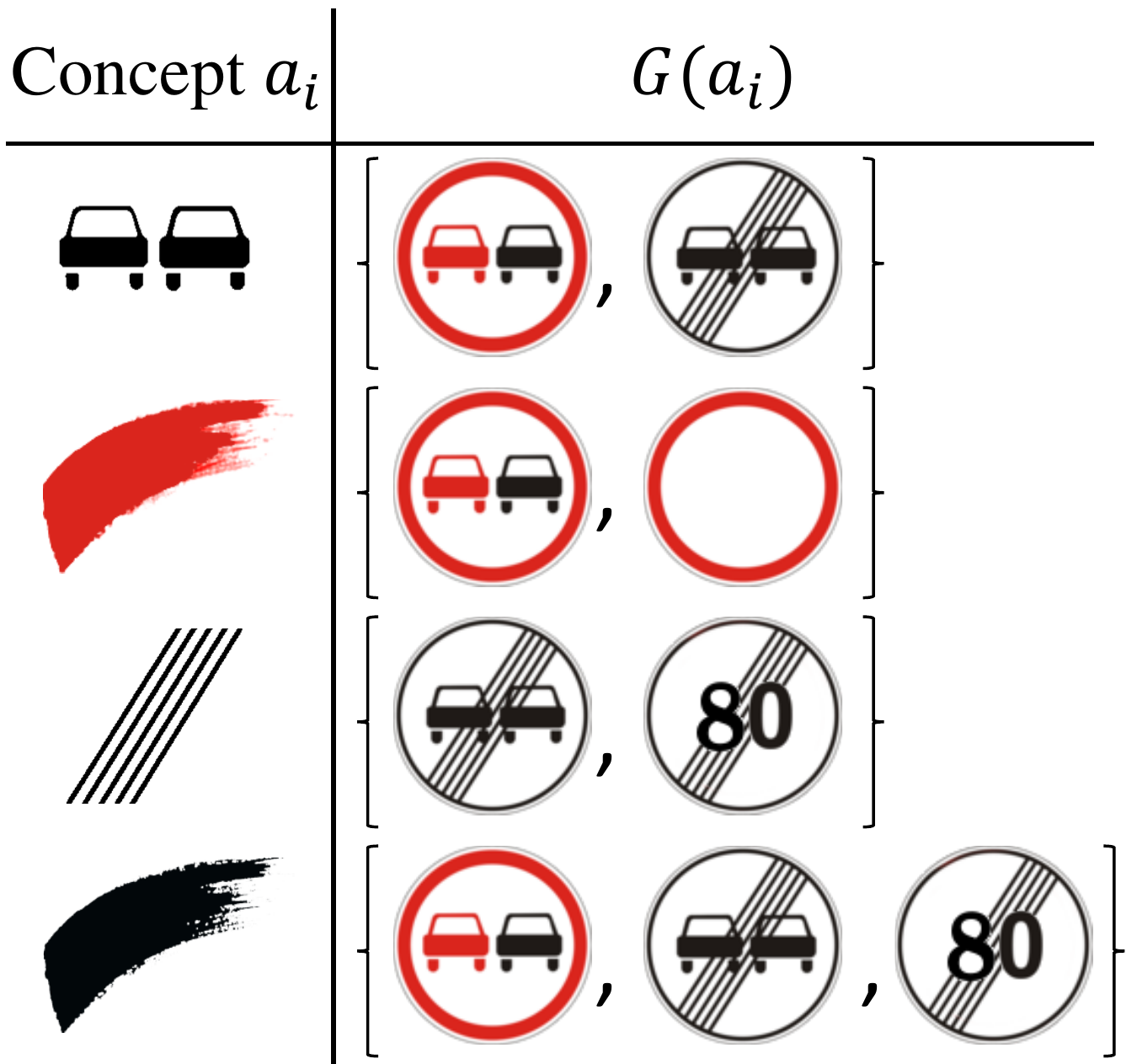}}
    \end{minipage}
        \caption{Example of an image classification example with four classes.}
        \label{fig:four_cls_example}
	\end{center}
\end{figure}

To score the concepts based on their contribution to the error detection performance of the SE-DL system, we first need to consider which classes are associated with each concept. Therefore, for each concept $a_i$, we define $G(a_i)$ as follows:

\begin{definition}\label{def:concept_G}
\textbf{Associated Classes Function $G(a_i)$:} For a concept $a_i\in\mathbb{A}:\{a_1,\dots,a_M\}$, we define the associated classes function $G(a_i): \mathbb{A}\rightarrow \mathcal{P}(\mathbb{C})$, where $\mathcal{P}(\mathbb{C})$ is the power set of classes $\mathbb{C}$. The $G(a_i):=\big\{C_j\in\mathbb{C} | a_i \in E(C_j)\big\}$ denotes the set of classes which concept ``$a_i$'' explains. In other words, $G(a_i)$ is the set of all classes that are associated with concept $a_i$.
\end{definition}

Figure~\ref{fig:four_cls_example} shows an example of an image classification problem with four classes based on the GTSRB~\cite{Houben-IJCNN-2013} dataset. Figures~\ref{fig:four_cls_class_example} and~\ref{fig:four_cls_concept_example} show the classes and concepts, respectively, given from the dataset of the classification problem. Also, Figure~\ref{fig:four_cls_g_example} illustrates the associated classes with each concept. For the concept ``Two cars'', we have $G($``Two cars''$)=\big\{$~``No passing'', ``End of no passing zone''~$\big\}$. Furthermore, we denote $\overline{G(\text{``Two cars''})}=\big\{$``Prohibited for all vehicles'', ``End of speed limit 80''$\big\}$ as the complement set of $G(\text{``Two cars''})$, which contains the classes that are not associated with the concept ``Two cars''.

\begin{figure}
	\begin{center}
        \subfloat[Detectable error using\\ $\:\:\:\:\:\:\:\:\:$Concept ``Two cars''
        \label{fig:det_err}]{\includegraphics[page={1}, width=0.38\columnwidth]{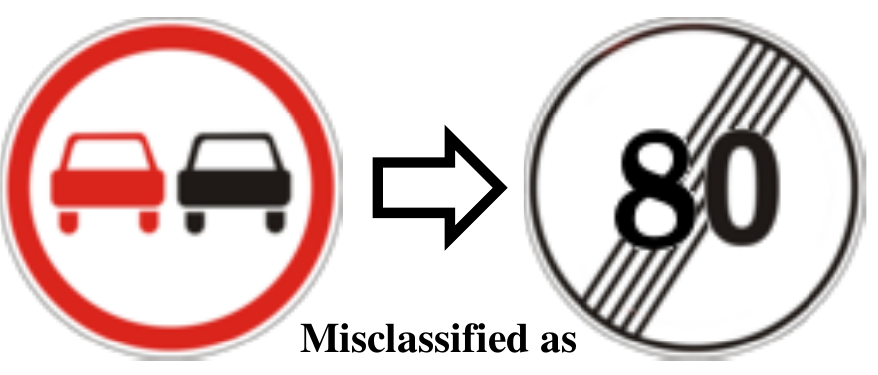}}
        \hspace{1cm}
        \subfloat[Non-detectable error using\\ Concept ``Two cars''
        \label{fig:nondet_err}]{\includegraphics[page={2}, width=0.38\columnwidth]{Img/det_nondet_err.pdf}}
        \caption{Examples of class prediction errors in the image classification of Figure~\ref{fig:four_cls_example}}
        \label{fig:det_nondet_err}
	\end{center}
\end{figure}

According to Definition~\ref{def:sed}, the presence or absence of a concept $a_i$ in a class explanation can be used to detect if the class is misclassified as certain other classes. Given the set of associated classes with a concept $a_i$, i.e., $G(a_i)$, we can determine which class prediction errors are detectable using that concept.
Figure~\ref{fig:det_nondet_err} shows examples of two class prediction errors based on the example of Figure~\ref{fig:four_cls_example}.  
The concept ``Two cars'' can be used to detect if an input image with ground truth class $C_x=$~``No passing'' is misclassified as the class $\hat{C}=$~``End of speed limit 80'' (Figure~\ref{fig:det_err}). The reason is that the ground truth class ``No passing'' $\in$ $G($``Two cars''$)$ and the predicted class ``End of speed limit 80'' $\in$ $\overline{G(\text {``Two cars''})}$, i.e., ``Two cars''$\not\in E($``End of speed limit 80''$)$. Therefore, the absence of the concept ``Two cars'' in the explanation of the ``End of speed limit 80'' leads to detecting such class prediction error. 

In contrast, in the example shown in Figure~\ref{fig:nondet_err}, the concept ``Two cars'' cannot be used to detect if the input image with ground truth class $C_x=$~``No passing'' is misclassified as the class $\hat{C}=$~``End of no passing zone''. The reason is that the ground truth class ``No passing'' $\in$ $G($``Two cars''$)$ and the predicted class~``End of no passing zone'' $\in$ $G($``Two cars''$)$, i.e., ``Two cars'' $\in$ $E($``No passing``$)$ and ``Two cars`` $\in$ $E($``End of no passing zone''$)$. Therefore, the concept is present in both the predicted explanation of the ground truth and the explanation of the erroneously predicted class and cannot be used to detect the class prediction error.

\begin{center}
\begin{algorithm}[t]
\small
\SetKwComment{Comment}{//\ }{}
\SetCommentSty{normal}
    \caption{Computing concept scores.}
    \label{alg:concept_selection}
    \textbf{Input:} Set of concept $\mathbb{A}:\{a_1,\dots,a_M\}$, set of classes $\mathbb{C}:\{C_1,\dots,C_N\}$\\
    \textbf{Output:} Overall scores $S_{ov}$\\
    
    \For{$\forall a_i\in \mathbb{A}$}{
    $S_{imp}(a_i)\leftarrow$ Using Equation~\ref{eq:importance_score} to compute {importance score} for the concept $a_i$\\
    $S_{sim}(a_i) \leftarrow $ Using Equation~\ref{eq:similarity_score} to compute {similarity score} for the concept $a_i$\\
    \Comment{Compute {overall score} using Equation~\ref{eq:similarity_score}:}
    $S_{ov} (a_i) \leftarrow  \frac{\alpha S_{imp}(a_i)}{\beta S_{sim}(a_i)}$\\
    }
    \textbf{Return} $S_{ov}$\\
\end{algorithm}
\end{center}

Based on the example above, we observe that a concept $a_i$ can be used to detect certain class prediction errors according to both sets $G(a_i)$ and $\overline{G(a_i)}$. We define the set of detectable class prediction errors using concept $a_i$ as follows:

\begin{definition}\label{def:cls_det_err}
\textbf{Detectable Class Prediction Errors Using Concept $a_i$, $DetErr(a_i)$:}  A concept $a_i$ can be used to detect a certain set of class prediction errors based on Equation~\ref{eq:deterr}:
\begin{equation}
\small
    \begin{aligned}~\label{eq:deterr}
    DetErr(a_i) = &\Big\{(C_j, C_k)\big| \big(C_j\in G(a_i)\:\:\:\wedge\:\:\:  C_k\in \overline{G(a_i)}\big)\:\:\vee\:\: \\
    &\big(C_j\in \overline{G(a_i)}\:\:\:\wedge\:\:\: C_k\in G(a_i)\big) \Big\}
\end{aligned}
\end{equation}

where $\overline{G(a_i)}=\mathbb{C}-G(a_i)$ and $(C_j,C_k)$ denotes that $C_j$ is misclassified as $C_k$.
\end{definition}

The goal of the concept selection is to score all concepts in $\mathbb{A}$ and select a subset of concepts $\mathbb{A}^*\subseteq\mathbb{A}$ based on the contribution of each concept to the error detection performance of the SE-DL system. Indeed, the set of detectable class prediction errors for concept $a_i$ can be used to quantify the contribution of each concept to the error detection performance.


Given a set of concepts, we quantify the overall contribution of each concept to the error detection performance of the SE-DL system using an \textit{overall} score. The overall score combines \textit{importance} and \textit{similarity} scores of each concept (shown in Algorithm~\ref{alg:concept_selection}).  We define the importance score to favor concepts that can be used to detect a larger number of class prediction errors and, therefore, have a higher contribution to the error detection performance of the system. Additionally, we define the similarity score to penalize concepts that detect class prediction errors similar to those of other concepts and, therefore, contribute less to the error detection performance of the system. 

\subsection{Concept Importance Score}\label{sec:importance_concept}~
The total number of detectable class prediction errors using a concept $a_i$ (Definition~\ref{def:cls_det_err}) represents the contribution of that concept to the error detection performance of the proposed SE-DL system. We use the cardinality of the set of detectable errors using concept $a_i$, i.e., $|DetErr(a_i)|$ to compute the importance score of that concept.

Consider a single-class image classification problem with $N$ classes in the dataset. Based on Equation~\ref{eq:deterr}, the total number of detectable class prediction errors using concept $a_i$ is computed as follows:
\begin{equation}\label{eq:det_err_size1}
\small
    |DetErr(a_i)| = \big(|G(a_i)|\times|\overline{G(a_i)}|\big) + \big(|\overline{G(a_i)}|\times|G(a_i)|\big)
\end{equation}
where $\overline{G(a_i)}$ denotes the complement set of $G(a_i)$, i.e., $\overline{G(a_i)}=\mathbb{C}-G(a_i)$. Since $N=|\mathbb{C}|$ is the number of classes in the dataset, we substitute  $|\overline{G(a_i)}| = N-|G(a_i)|$ in Equation~\ref{eq:det_err_size1} and re-write it as:
\begin{equation}\label{eq:det_err_size2}
\small
    |DetErr(a_i)| = 2|G(a_i)|\big(N-|G(a_i)|\big)
\end{equation}

Assume that each class has a uniform chance of misclassifying as $N-1$ other classes. The total number of class prediction errors is $N(N-1)$, representing the number of possible ways in which $N$ classes can be misclassified as any $N-1$ other classes. 
We define the importance score of the concept $a_i$ as follows:

\begin{definition}\label{def:importance_score}\textbf{Importance Score of a Concept $S_{imp}(a_i)$}~ is the ratio of the number of detectable class prediction errors using the concept $a_i$ to the total number of possible class prediction errors. We compute the $S_{imp}(a_i)$ as Equation~\ref{eq:importance_score}:
\begin{equation}\label{eq:importance_score}
\small
\begin{aligned}
   S_{imp}(a_i) &= \frac{|DetErr(a_i)|}{\text{Total number of possible errors}}\\
   &=\frac{2\:|G(a_i)|\: \big(N-|G(a_i)|\big)}{N\:(N-1)}
\end{aligned}
\end{equation}
\end{definition}

\subsection{Concept Similarity Score}\label{sec:similarity_Concepts}~
We define the similarity score to penalize concepts that contribute less to the error detection performance of the proposed SE-DL system by detecting a set of class prediction errors similar to those of other concepts. We use the set of detectable class prediction errors using concept $a_i$ (Definition~\ref{def:cls_det_err}) to compute the concept similarity score.

Consider concept $a_i$ that detects certain class prediction errors. The use of another concept $a_j$ that detects a similar set of class prediction errors to that of $a_i$ does not contribute more to the SE-DL system's overall set of detectable class prediction errors. In the example shown in Figure~\ref{fig:four_cls_example}, we observe that the set of detectable class prediction errors using concept ``Color red'' is the same as that of the concept ``Parallel tilted lines'', i.e., $DetErr\big(\text{``Color red''}\big) = DetErr\big(\text{``Parallel tilted lines''}\big)$. Therefore, using both of these concepts does not contribute more to the error detection performance than using only one of them.

Consequently, we quantify the similarity between concepts using \textit{Jaccard similarity index}~\cite{niwattanakul2013using}. The Jaccard similarity index is widely adopted in deep learning applications for calculating the similarity between set samples~\cite{besta2020communication, zafar2019dlime, wu2017extracting}. Besta~\textit{et~al.}~\cite{besta2020communication} use the Jaccard similarity index to compute the similarity among pairs of large datasets in a data sequencing problem. Wu~\textit{et~al.}~\cite{wu2017extracting} use the Jaccard similarity index to calculate the similarity of sets of words in a document for a clustering problem. Here, we use the Jaccard similarity index to identify the similarity between a pair of concepts in terms of the class prediction errors they detect. The similarity between two sets $A$ and $B$ can be measured using the Jaccard similarity index as follows~\cite{niwattanakul2013using}:
\begin{equation}\label{eq:jaccard}\small
    \text{Jaccard}(A,B) = \frac{|A\bigcap B|}{|A\bigcup B|} = \frac{|A\bigcap B|}{|A| +|B|- |A\bigcap B|}
\end{equation}

The Jaccard similarity index is a positive number that ranges between 0, when $A\bigcap B=\{\}$, and 1, when $A\bigcap B = A\bigcup B$. Using Equation~\ref{eq:jaccard}, we compute the pairwise Jaccard similarity between concepts $a_i$ and $a_j$ in terms of their sets of detectable class prediction errors as follows:
\begin{equation}\label{eq:pair_similarity}\small
    \text{Jaccard}\big(DetErr(a_i),DetErr(a_j)\big) = \frac{\big|DetErr(a_i)\bigcap DetErr(a_j)\big|}{\big|DetErr(a_i)\bigcup DetErr(a_j)\big|}
\end{equation}

Based on the pairwise Jaccard similarity of concepts in Equation~\ref{eq:pair_similarity}, we define the similarity score of the concept $a_i$ as follows:
\begin{definition}\label{def:similarity_score}
\textbf{Similarity Score of a Concept $S_{sim}(a_i)$} is computed by taking the average pairwise Jaccard similarity of the concept $a_i$ and other concepts $\forall a_j \in \mathbb{A}, j\neq i$ as shown in Equation~\ref{eq:similarity_score}:
\begin{equation}\label{eq:similarity_score}
\small
    S_{sim}(a_i) = \frac{1}{|\mathbb{A}|-1}\mathlarger{\sum}_{\forall a_j\:\in\:\mathbb{A},\:j\neq i} \text{Jaccard}\big(DetErr(a_i),DetErr(a_j)\big)
\end{equation}
where $|\mathbb{A}|$ denotes the total number of concepts identified in the dataset, and the term $\frac{1}{|\mathbb{A}|-1}$ in Equation~\ref{eq:similarity_score} normalizes the similarity score between 0 and 1. 
\end{definition}
Assuming that any concept $a_i\in \mathbb{A}$ explains at least one class i.e., $G(a_i)\neq\{\}$, and not all the classes, i.e., $\overline{G(a_i)}\neq\{\}$, by contradiction we can prove that the similarity score of concept $a_i$ is always non-zero, i.e., $S_{sim}(a_i) \neq 0$.

\subsection{Concept Overall Score}\label{sec:overall_score_concepts}~
The overall score of a concept $a_i\in \mathbb{A}$ combines the importance and the similarity scores and determines the overall contribution of the concept to the error detection performance of the proposed SE-DL system. A concept has a higher overall score when it has a higher importance score and a lower similarity score.

We define the overall score $S_{ov}(a_i)$ of the concept $a_i$ as the ratio of the concept importance score over the concept similarity score.
Equation~\ref{eq:overall_score} shows the concept overall score:

\begin{equation}
\label{eq:overall_score}
\small
S_{ov}(a_i) = \frac{\alpha S_{imp}(a_i)}{\beta S_{sim}(a_i)}
\end{equation}
where $S_{imp}(a_i)$ and $S_{sim}(a_i)$ are the importance and similarity scores of the concept $a_i$ calculated from Equations~\ref{eq:importance_score} and~\ref{eq:similarity_score}, respectively.  The $\alpha$ and $\beta$ are normalization coefficients to produce comparable importance and similarity scores. We normalize the importance score for the concept $a_i$ by dividing it over the sum of the importance scores of all concepts. Likewise, we normalize the similarity score for the concept $a_i$ by dividing it over the sum of the similarity scores of all concepts. Hence, we have:
\begin{equation}
\begin{aligned}
    \alpha = \frac{1}{\sum_{\forall j} S_{imp}(a_j)}, \:\:\:  \beta = \frac{1}{ \sum_{\forall j} S_{sim}(a_j)}
\end{aligned}    
\end{equation}

%% file: 06_eval_methodology.tex
\section{Evaluation Methodology}\label{sec:err_detection_scheme}~
In this section, we discuss the methodology to evaluate the error detection performance (Definition~\ref{def:perf_err_det}) of the proposed self-explainable deep learning~(SE-DL) system (Subsection~\ref{sec:prop_self_exp_sys}) and compare it with the existing ensemble technique~\cite{tramer2018ensemble, sen2020empir, wei2020cross}. Additionally, we discuss the methodology to evaluate the error detection performance of the proposed SE-DL system when it is integrated into an ensemble error detection scheme.

\begin{figure}
\centering
\hspace{-1.2cm}
	\subfloat[R1
 \label{fig:r1_emodel}]{\includegraphics[page={1}, width=0.35\columnwidth]{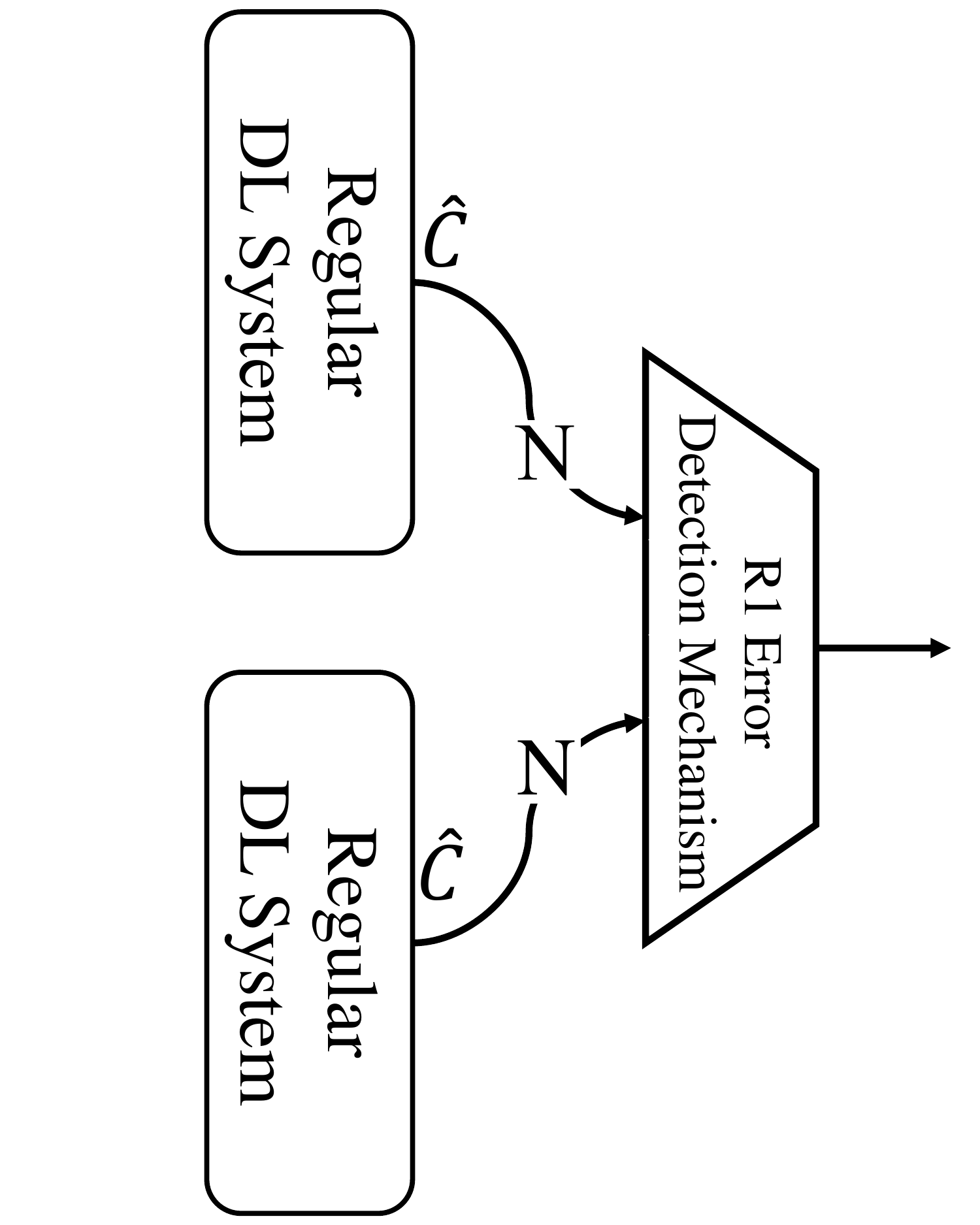}}
 \subfloat[SE
 \label{fig:se_emodel}]{\includegraphics[page={2}, width=0.35\columnwidth]{Img/Error_Detection_Models.pdf}}
  \subfloat[SE+R1
  \label{fig:r1_se_emodel}]{\includegraphics[page={3}, width=0.35\columnwidth]{Img/Error_Detection_Models.pdf}}
 \caption{Different error detection schemes. 
 }
 \label{fig:error_detection_scheme}
\end{figure}

Figure~\ref{fig:error_detection_scheme} shows the three error detection schemes that we evaluate. The error detection schemes consist of deep learning~(DL) systems designed for a single-class image classification problem with $N$ classes in the dataset. There are two different types of DL systems for a single-class image classification problem: \textit{regular} and \textit{self-explainable}. A regular DL system generates $N$ output predictions that indicate the predicted class of the input image. A \textit{self-explainable}  DL system  generates $N+M^*$ output predictions, which include both $N$  output predictions as the predicted class and $M^* = |\mathbb{A}^*|$ concepts as the predicted explanation $\hat{E}$.

\subsection{R1 Error Detection Scheme}\label{sec:scheme_r1}~
The R1 error detection scheme is based on the existing ensemble techniques~\cite{tramer2018ensemble, sen2020empir, wei2020cross}. Figure~\ref{fig:r1_emodel} illustrates the R1 error detection scheme which consists of two regular DL systems trained on different splits of the training dataset. Both DL systems take the same image as their inputs, and each independently generates $N$ output predictions, indicating their predicted classes of the input image $\hat{C}$. 
    
In the R1 scheme, the error detection mechanism takes the predicted classes from both DL systems and compares them. Specifically, an error is detected when the two predicted classes do not match.

\subsection{SE (\textit{Self-Explainable}) Error Detection Scheme}\label{sec:scheme_se}~
The SE error detection scheme is based on the proposed SE-DL image classifier system in Subsection~\ref{sec:prop_self_exp_sys}. Figure~\ref{fig:se_emodel} shows the SE error detection scheme, which consists of a single SE-DL system. The SE-DL system takes in the input image and generates $N$ output predictions indicating the predicted class of the input image $\hat{C}$, as well as $M^*$ concepts as the predicted explanation $\hat{E}$ of the class of the input image.

In SE scheme, the errors are detected by the proposed self-error detection mechanism (Subsection~\ref{sec:prop_self_err_det_mech}). This mechanism takes in the predicted class and the predicted explanation from the SE-DL system. An error is detected when the ground truth explanation of the predicted class $E(\hat{C})$ does not match  the predicted explanation $\hat{E}$, i.e., $E(\hat{C})\neq\hat{E}$.

\subsection{SE+R1 Error Detection Scheme}\label{sec:scheme_ser1}~
The SE+R1 error detection scheme enhances the SE scheme by integrating it into an ensemble technique. Specifically, the SE+R1 error detection scheme consists of a single SE-DL system and an additional regular DL system. Both the SE-DL and the regular DL systems take in the same input image. The SE-DL system generates $N$ output predictions indicating the predicted class of the input image $\hat{C}$, as well as $M^*$ concepts as the predicted explanation $\hat{E}$ of the class of the input image. Additionally, the regular DL system independently generates $N$ output predictions indicating the predicted class of the input image $\hat{C}$.

In the SE+R1 scheme, the errors are detected by a self-error detection mechanism (Subsection~\ref{sec:prop_self_err_det_mech}), as well as by comparing the predicted classes of both the SE-DL and regular DL systems. Specifically, an error is detected when the ground truth explanation of the predicted class $E(\hat{C})$ does not match the predicted explanation $\hat{E}$, or the predicted classes $\hat{C}$ from both SE-DL and regular DL systems do not match.

%% file: 07_results.tex
\section{Results}\label{sec:res}~
In this section, we describe the results of evaluating the proposed methodology in Section~\ref{sec:proposed_method}.

\textbf{Dataset Specifications.}\label{sec:res_prop_method}~
We use the German Traffic Sign Recognition Benchmark (GTSRB) dataset~\cite{Houben-IJCNN-2013} to train DL image classifier systems that classify road traffic signs. The GTSRB dataset contains 43 classes of different traffic signs, including data from different weather conditions, light illumination, occlusions, etc. The GTSRB dataset includes 39,209 images and their ground truth classes which we divide into training (70\%) and verification (30\%) datasets. The GTSRB dataset also includes 12,630 images and their ground truth classes as the validation dataset.

\textbf{Machine Specifications.}~
We use a DGX~A100 GPU system from Nvidia GPU Cloud (NGC) to implement the proposed methodology. The DGX system contains 8~A100 Tensor Core GPUs with a total 40~GB GPU Memory, and 240~CPU cores with 976~GB system memory. The algorithm is implemented using PyTorch 1.7.0 machine learning framework.

\textbf{DL Architecture Specifications.}~
We employ \textit{transfer learning} method~\cite{dai2007boosting} to reuse the optimized weights from a pre-trained Deep Learning~(DL) system such as AlexNet~\cite{krizhevsky2017imagenet}. AlexNet architecture is widely used for image classification tasks~\cite{gao2018object, melotti2018cnn, zhao2018dha} to develop object classification systems and classify images obtained from vision data in an autonomous vehicle environment~\cite{gao2018object}. We use AlexNet DL system that was previously trained on ImageNet dataset with $N=1000$ classes~\cite{krizhevsky2017imagenet, krizhevsky2012imagenet}.  
However, the number of classes in the ImageNet dataset differs from the proposed SE-DL image classifier system built for GTSRB. Particularly, there are $1000$ classes in ImageNet instead of $43$ classes in the SE-DL image classifier. To address this problem, we replace the output layer of the pre-trained DL system with a new Softmax output layer and re-train the resulting model with the GTSRB dataset. 

Our AlexNet-based DL architecture consists of five convolutional layers with Relu activation functions, four pooling layers, two fully-connected layers with Relu activation functions, two dropouts for the fully-connected layers and a  fully-connected softmax output layer. For the GTSRB dataset, the softmax output layer for the regular classifier consists of 43 neurons indicating the predicted class of the input image. Also, the softmax output layer for the SE-DL classifier consist of $43+M^*$ neurons indicating the predicted class along with the predicted explanation. 

In Subsection~\ref{sec:res_comp_comp}, we analyze the computational complexity and memory size of the regular and the proposed SE-DL classifiers. In Subsection~\ref{sec:res_concept_sel}, we present the results of the concept selection methodology and the overall scores of the concepts for the GTSRB dataset. Finally, in Subsection~\ref{sec:res_eval_err_det_scheme}, we elaborate the results of evaluating the error detection performance of the different error detection schemes of Section~\ref{sec:err_detection_scheme}. 

\input{Tables/tab_dl_net_paper}
\subsection{Analysis of Computational Complexity and Memory Size}\label{sec:res_comp_comp}
Computational complexity of a DL system can be analyzed using computational load (number of multiplication operations) and memory occupation (number of weights stored in the memory)~\cite{alippi2018moving, alom2018history}. 

Table~\ref{tab:dl_net_paper} entails the number of multiplication operations (computed based on~\cite{alom2018history}) and the number of weights for both the SE-DL and the regular DL classifier systems. Specifically, the total number of multiplication operations of the proposed SE-DL system is only $0.007\%$ higher compared to the regular classifier. As a result, the computational complexity of both classifiers is comparable. Moreover, the number of weights determines the amount of memory required to store the DL system parameters. Compared to the regular classifier, the proposed SE-DL classifier requires only $0.086\%$ more memory in total. Hence, the memory size of both classifiers is comparable.

Additionally, we  analyze the computational complexity and memory size of error detection schemes discussed in Section~\ref{sec:err_detection_scheme}. Recall that R1 consists of two regular classifiers, and SE+R1 consists of one regular and one SE-DL classifier. As a result, the R1 and the SE+R1 error detection schemes have comparable computational complexity and memory size. On the other hand, the SE scheme consists of only a single SE-DL classifier. Thus, it has approximately half of the computational complexity of the other two schemes.

\begin{figure}
\centering
	\subfloat[Classes in the GTSRB~\cite{Houben-IJCNN-2013} dataset.
 \label{fig:gtsrb_data}]{\includegraphics[width=0.77\columnwidth]{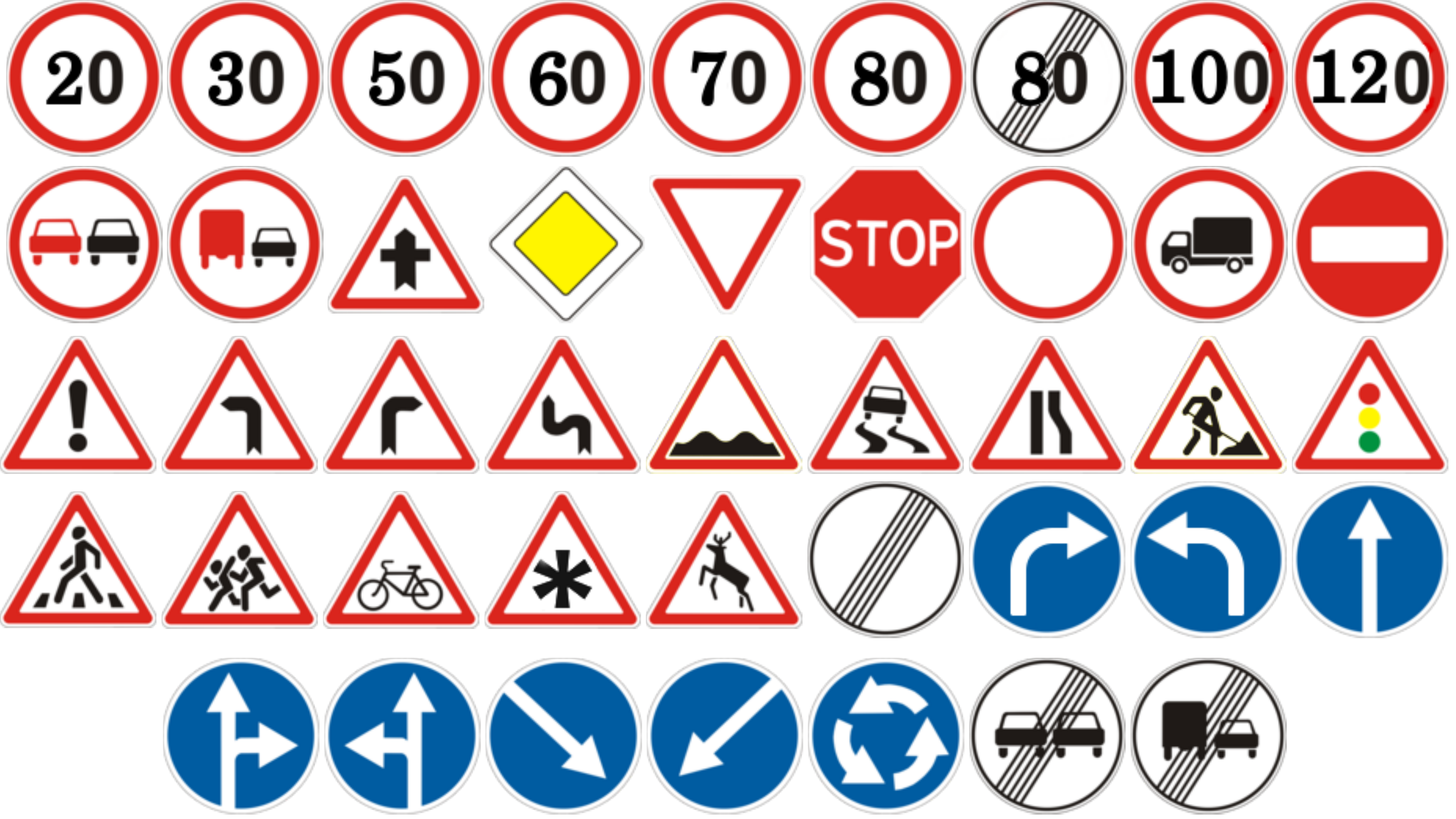}}
 \hspace{0.77\columnwidth}\\
 \subfloat[Concepts identified in the GTSRB~\cite{Houben-IJCNN-2013} dataset.
 \label{fig:gtsrb_concepts}]{\includegraphics[width=0.6\columnwidth]{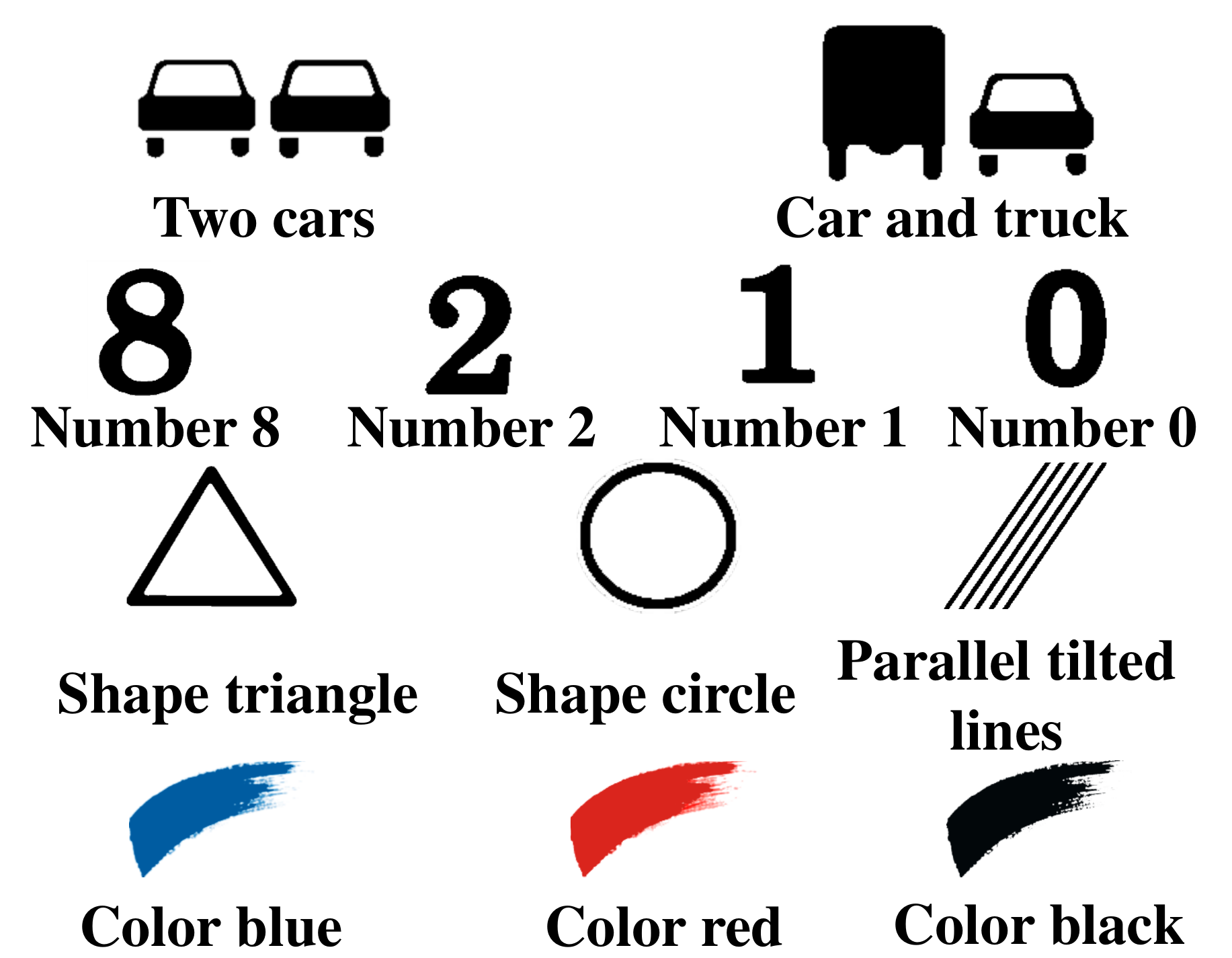}}
 \caption{German Traffic Sign Recognition Benchmark (GTSRB) dataset~\cite{Houben-IJCNN-2013}.}
 \label{fig:gtsrb}
\end{figure}

\subsection{Results of Concept Selection Methodology}\label{sec:res_concept_sel}~
In this subsection, we present the results of the concept selection methodology presented in Section~\ref{sec:concept-selection}. 
Figure~\ref{fig:gtsrb_data} shows 43 classes of traffic signs that exist in the GTSRB dataset. 
Based on Definition~\ref{def:concept}, we identify a set of concepts $\mathbb{A}$ from the classification problem domain that explains at least two classes and uses the concepts for generating explanations (shown in Figure~\ref{fig:gtsrb_concepts}). We develop a SE-DL system with $N=43$ classes and $M^*$ highest scored concepts selected using  Algorithm~\ref{alg:concept_selection}.

\begin{figure}
    \centering
		\includegraphics[page={1}, width=0.95\columnwidth]{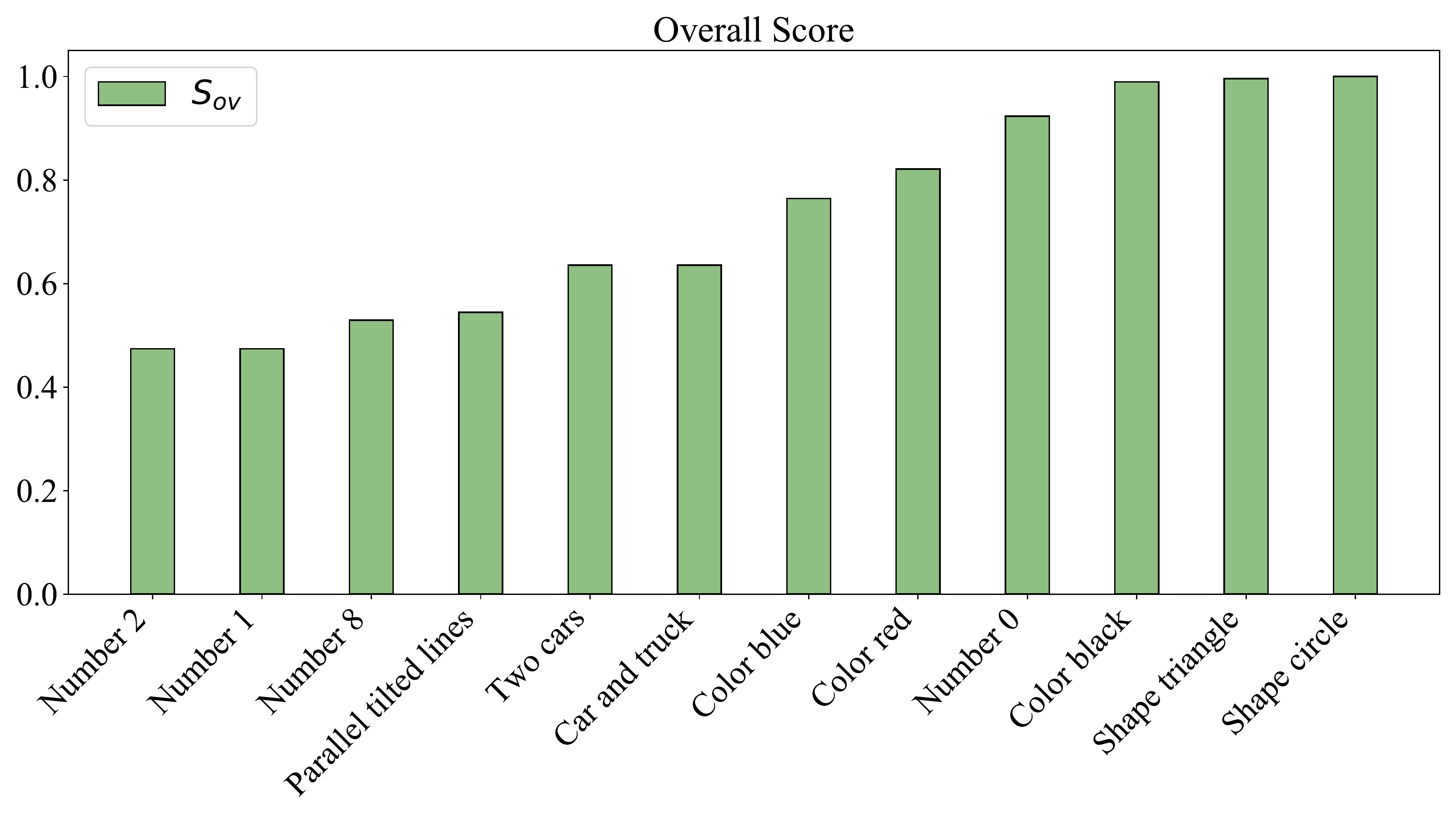}
		\caption{Overall score $S_{ov}$ of the concepts in the GTSRB dataset.}
    \label{fig:overall_scores}
\end{figure}

\begin{figure}
    \centering
		\includegraphics[page={2}, width=0.95\columnwidth]{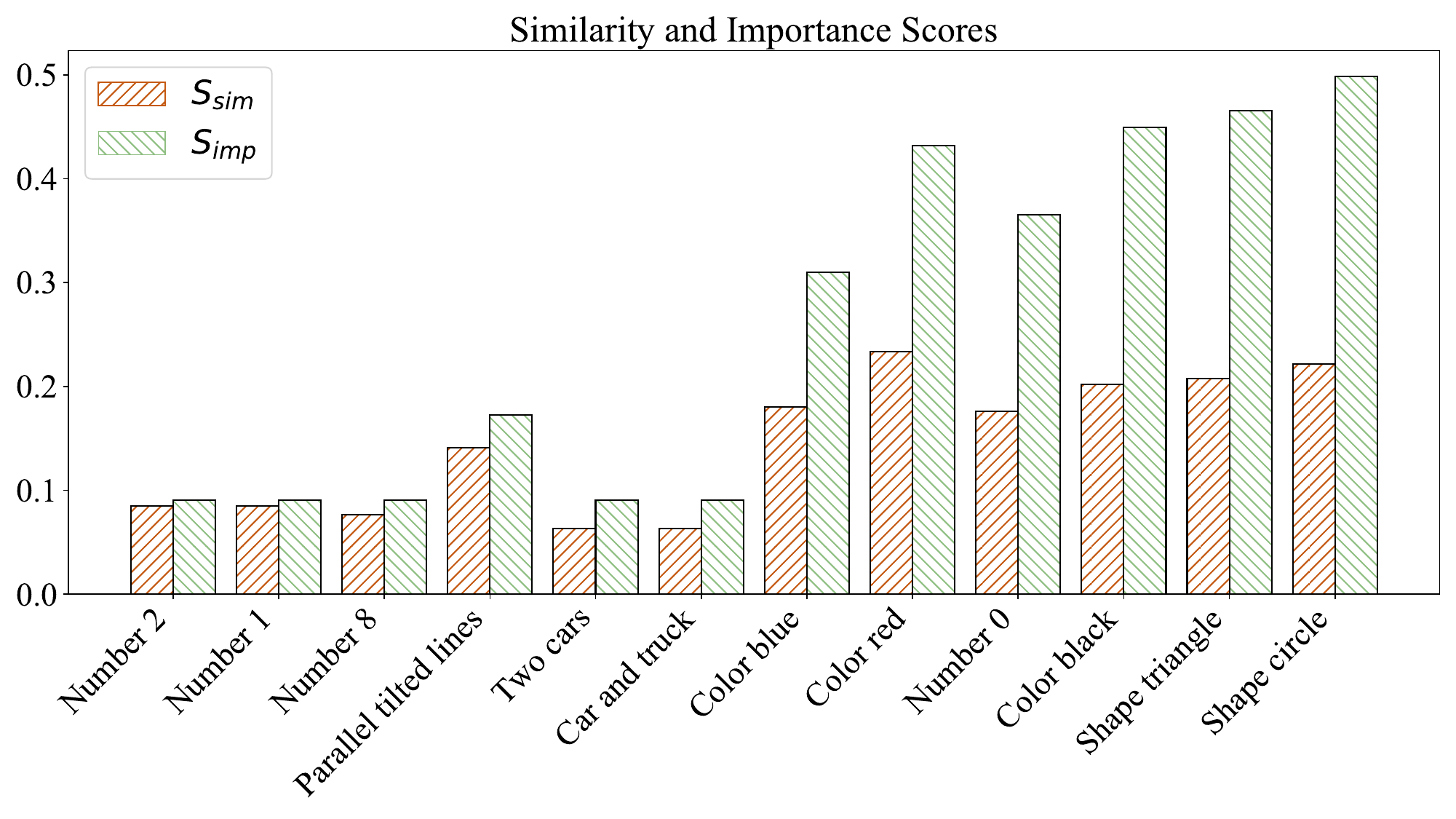}
		\caption{Importance score $S_{imp}$ and similarity score $S_{sim}$ for the GTSRB dataset.}
    \label{fig:s_imp_s_sim}
\end{figure}

\textbf{Evaluating Concept Scores.}\label{sec:res_sel_imp_concept}~
Figure~\ref{fig:overall_scores} shows the overall scores $S_{ov}$ for each concept computed using Algorithm~\ref{alg:concept_selection} and normalized to $[0,1]$. Also, Figure~\ref{fig:s_imp_s_sim} shows the importance score $S_{imp}$ and the similarity score $S_{sim}$ of the concepts for the GTSRB dataset in the order they appear in Figure~\ref{fig:overall_scores}. A concept has a higher overall score if it has a higher importance score and a lower similarity score. According to Figure~\ref{fig:s_imp_s_sim}, the ``Shape circle'' has a high importance score with a relatively smaller similarity score. Hence, the ``Shape circle'' has the highest overall score as shown in Figure~\ref{fig:overall_scores}. In contrast, the ``Number 2'' and ``Number 1'' have the least overall scores due to their low importance scores relative to their high similarity scores.


\subsection{Results of Evaluating Error Detection Performance}\label{sec:res_eval_err_det_scheme}~
In this subsection, we demonstrate the results of evaluating the error detection performance (Definition~\ref{def:perf_err_det}) of the error detection schemes using adversarial examples. We compare the performance of the two proposed error detection schemes based on the SE-DL image classifier, i.e., SE and SE+R1, with the existing ensemble technique~\cite{tramer2018ensemble, sen2020empir, wei2020cross}, i.e., R1.

\textbf{Generating Adversarial Examples.}\label{sec:res_met_eval_rob}~
We generate adversarial examples to analyze the performance of the error detection schemes presented in Section~\ref{sec:err_detection_scheme}. 
Adversarial examples alter the prediction of the DL image classification system by generating perturbed inputs. Particularly, we use Fast Gradient Sign Method~(FGSM)~\cite{goodfellow2015explaining}, due to its simple yet effective implementation, to generate adversarial examples. Moreover, FGSM has been used by other research works to evaluate the error detection performance of the DL systems against adversarial perturbations~\cite{ross2018improving, wang2019convergence, li2020implicit}. 
The FGSM perturbation typically varies between $[0-0.1]$ to create subtle perturbations~\cite{pang2019improving, goodfellow2015explaining}. We try three FGSM perturbation rates in the range of $[0-0.15]$: 0.05, 0.1, and 0.15.

\begin{figure}
\centering
	\subfloat[$\epsilon = 0$ \\(``Speed limit 30'')
 \label{fig:ex_fgsm_orig}]{\includegraphics[width=0.25\columnwidth]{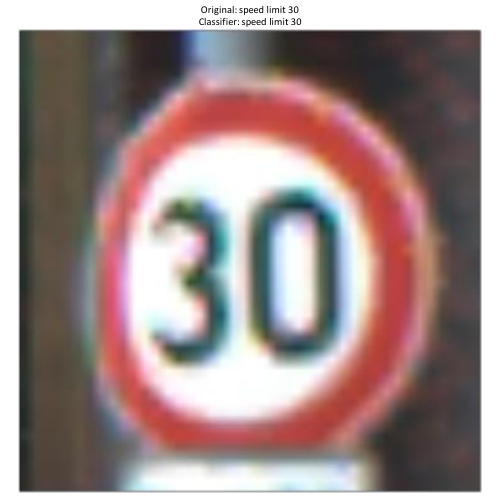}}
 \subfloat[$\epsilon = 0.05$ \\(``Danger'')
 \label{fig:ex_fgsm_at.05}]{\includegraphics[width=0.25\columnwidth]{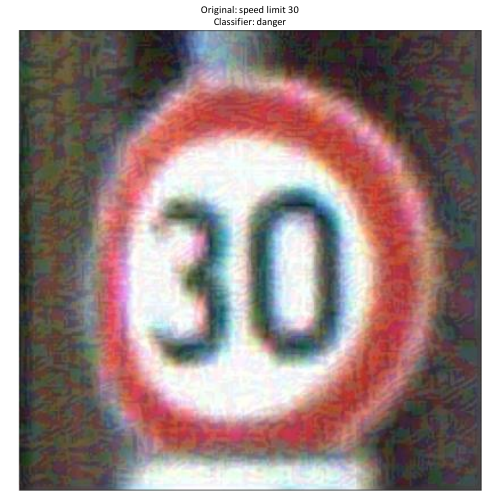}}
 \subfloat[$\epsilon = 0.1$ \\(``Danger'')
 \label{fig:ex_fgsm_at.1}]{\includegraphics[width=0.25\columnwidth]{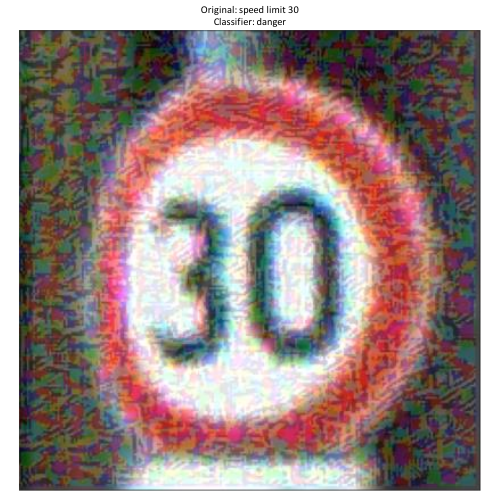}}
 \subfloat[$\epsilon = 0.15$ \\(``Double curve'')
 \label{fig:ex_fgsm_at.15}]{\includegraphics[width=0.25\columnwidth]{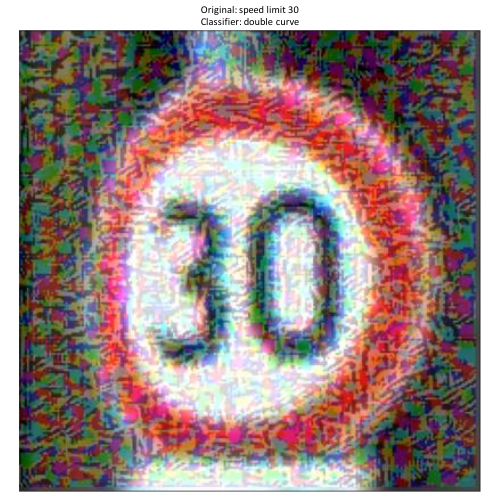}}
 \caption{Examples of FGSM adversarial example attacks with different perturbation rates.}
 \label{fig:ex_fgsm}
 
\end{figure}

Figure~\ref{fig:ex_fgsm} shows adversarial examples with different perturbation rates on a traffic sign image of ``Speed Limit 30''. Figure~\ref{fig:ex_fgsm_orig} shows the original image, which is correctly classified by a regular image classifier DL system. Figures~\ref{fig:ex_fgsm_at.05}-\ref{fig:ex_fgsm_at.15} illustrate the perturbed images using perturbation rates of 0.05, 0.1, and 0.15. We observe that in all three cases, the DL system incorrectly predicts the traffic sign as ``Danger'' or ``Double Curve''. 

\begin{figure}
    \centering
    \includegraphics[width=0.65\columnwidth]{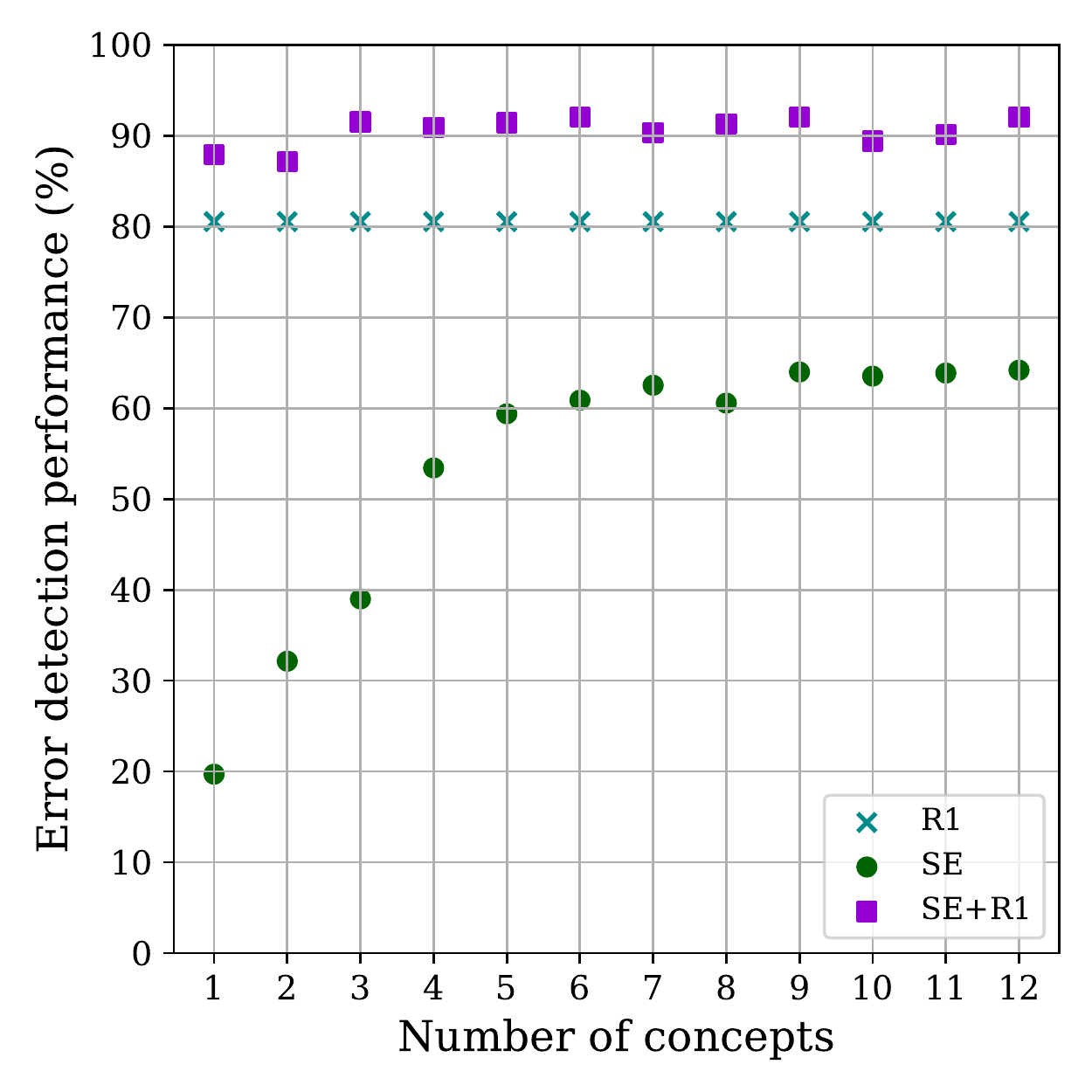}
    \caption{Impact of concept selection on error detection performance of different error detection schemes.}
    \label{fig:SI_eval}
\end{figure}

\textbf{Impact of Concept Selection on Error Detection Performance.}\label{sec:res_impact_concept_sel}~
Figure~\ref{fig:SI_eval} demonstrates how the proposed concept selection methodology can improve the error detection performance of the proposed self-error detection mechanism. The x-axis shows the number of concepts involved in the error detection schemes of SE and SE+R1, and the y-axis shows the error detection performance. The concepts involved in the error detection scheme are selected based on the overall score shown in Figure~\ref{fig:overall_scores}. 
Specifically, we first begin with incorporating the highest-scored concepts based on the overall score. We then gradually increase the number of concepts and measure the resulting error detection performance based on Definition~\ref{def:perf_err_det}. We use different FGSM perturbation rates ($\epsilon=$ 0.05, 0.1, and 0.15) to generate adversarial example input images and compute the average error detection performance for all perturbed images.


Figure~\ref{fig:SI_eval} shows that the SE error detection reaches 62.2\% error detection performance with only one SE-DL classifier, while R1 reaches 81.8\% with two regular classifiers. The SE+R1 scheme achieves a consistently higher error detection performance than the two other error detection schemes. The reason is that the SE+R1 scheme contains the improvements over both schemes by having an additional regular classifier, as well as a SE-DL classifier. Figure~\ref{fig:SI_eval} shows that the SE+R1 error detection scheme achieves error detection performance of higher than 92\% after adding the five highest scored concepts given in Figure~\ref{fig:overall_scores}. On the other hand, by including all twelve concepts, it attains an error detection performance of 92.1\%. This indicates that the minimal subset of concepts $\mathbb{A}^*$ for designing the SE-DL system contains the five highest scored concepts, i.e., $\mathbb{A}^*:\big\{$``Shape circle'', ``Shape triangle'', ``Color black'', ``Number 0'', ``Color red''$\big\}$. In contrast, the error detection performance of the R1 scheme remains constant at 80.5\% since it does not include any SE-DL system in its scheme.

\begin{figure}
    \centering
    \includegraphics[width=0.65\columnwidth]{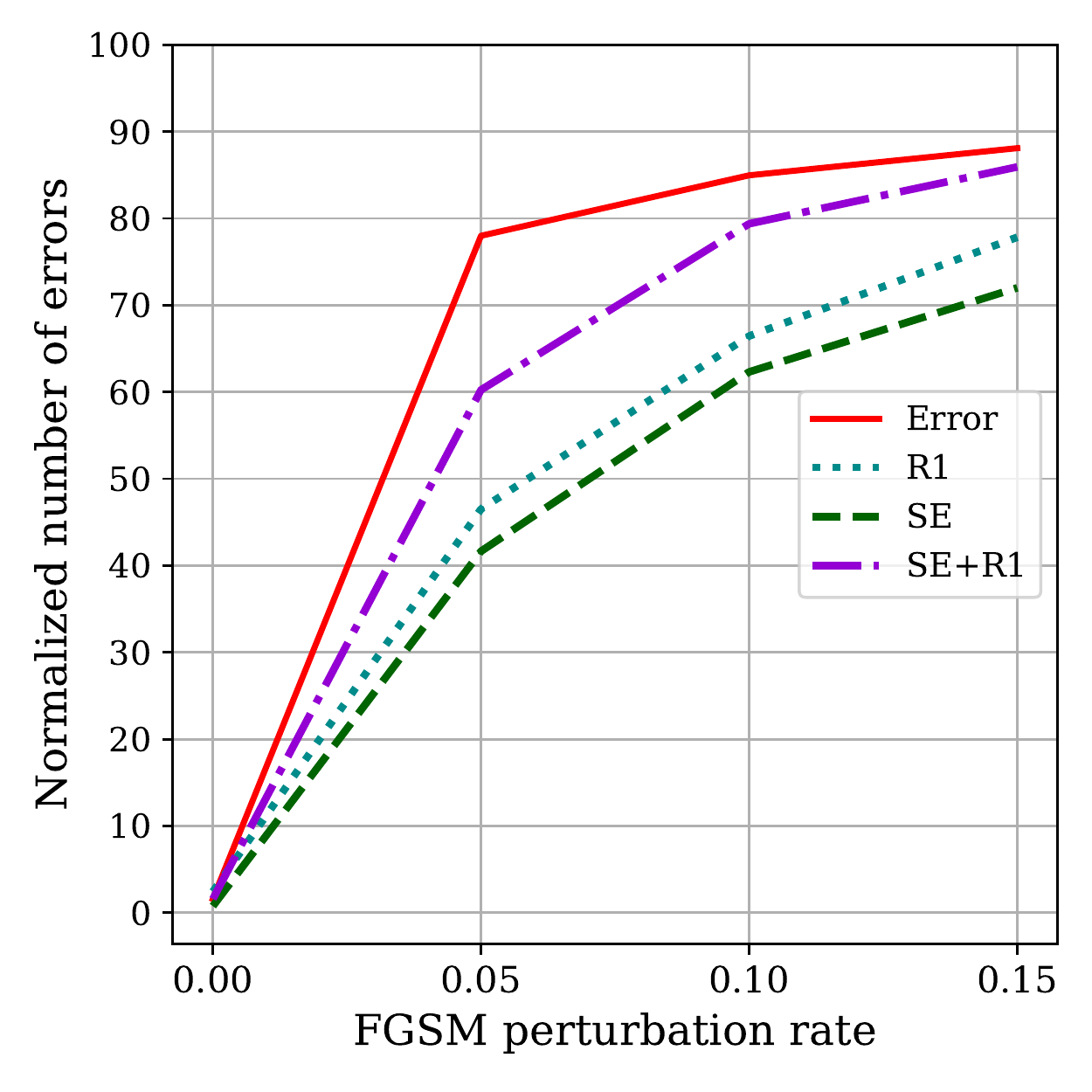}
    \caption{Comparing the number of detected errors for R1, SE and SE+R1 error detection schemes.}
    \label{fig:r1r2-3}
\end{figure}

\textbf{Error Detection Performance Comparison of Different  Schemes.}\label{sec:res-err_det_perf}~
We study the performance of different error detection schemes in the face of adversarial examples and illustrate the results in Figure~\ref{fig:r1r2-3}. The x-axis shows the FGSM perturbation ($\epsilon=$ 0.05, 0.1, and 0.15), and the y-axis shows the normalized number of errors out of 100 data samples. The higher the perturbation results in more prediction errors by the DL system.

Specifically, we compare the number of detected errors for the R1, SE, and SE+R1 error detection schemes in Figure~\ref{fig:r1r2-3}. The SE-DL systems used in SE and SE+R1 schemes are trained with the five highest scored concepts. Also, the R1 represents the existing ensemble technique using two regular classifiers.  
The solid red line shows the total number of prediction errors of the DL system. The dotted cyan, dashed green, and dot-dashed purple lines show the number of detected errors using R1, SE, and SE+R1 schemes, respectively. In this experiment, there is a total number of 88 errors out of each 100 data at the highest perturbation of $\epsilon = 0.15$. Using the number of detected errors shown in Figure~\ref{fig:r1r2-3}, we compute the performance of the error detection for each scheme.

As shown in Figure~\ref{fig:r1r2-3}, at the highest perturbation $\epsilon=0.15$, the R1 error detection scheme detects 77 errors out of a total number of 88 errors, while the proposed SE+R1 scheme detects 86 errors out of a total number of 88 errors. In other words, the proposed SE+R1 scheme consisting of one self-explainable and one regular classifier achieves an error detection performance of 97.7\% at the highest perturbation. In contrast, the baseline R1 scheme with two regular classifiers reaches an error detection performance of 87.5\% at comparable computational complexity and memory size as discussed in Subsection~\ref{sec:res_comp_comp}. Thus, achieving an error detection performance of SE+R1 at 97.7\% using the R1 scheme requires adding more regular classifiers, which incurs additional computational complexity and memory size compared to the proposed SE+R1 Scheme. 

Also, Figure~\ref{fig:r1r2-3} shows that the SE scheme detects 72 errors out of a total number of 88 errors. Therefore, the performance of error detection in the SE using a single self-explainable classifier is 81.8\%, which is comparable to the error detection performance of the R1 scheme (87.5\%) using two regular classifiers.

Based on the analysis from Subsection~\ref{sec:res_comp_comp}, the proposed R1 and SE+R1 schemes incur a similar computational complexity and memory size for their two classifiers. Moreover, the SE scheme incurs approximately half of the computational complexity and memory size of the former schemes, for it uses a single classifier. Hence, we conclude that the proposed SE+R1 exhibits a higher error detection performance compared to the existing R1 scheme at a comparable computational complexity and memory size. Additionally, the proposed SE scheme compared to the existing R1 scheme has approximately half of the computational complexity and memory size, with a slightly lower error detection performance at the highest perturbation.

%% file: Tables/tab_dl_net_paper.tex
\begin{smaller}

\begin{table}[t]
\centering
\smaller
\caption{Comparison of DL classifier architectures.}
\label{tab:dl_net_paper}
\begin{tabular}
{>{\raggedright\arraybackslash}p{0.25\columnwidth} >{\raggedright\arraybackslash}p{0.16\columnwidth}>{\raggedright\arraybackslash}p{0.11\columnwidth}>{\raggedright\arraybackslash}p{0.11\columnwidth}>{\raggedright\arraybackslash}p{0.12\columnwidth}}
\hline
 \textbf{DL Type} & \textbf{Softmax size} & \textbf{\#Neurons} & \textbf{\#Weights} & \textbf{\#Operations}  \\\hline\hline


The Proposed SE-DL & $43+M^*$& 
500,343   &   57,219,776   &   712,896,512\\

Regular DL\cite{krizhevsky2017imagenet} & $43$ & 
500,331   &    57,170,624  &    712,847,360 \\ \hline
\hline
\end{tabular}
\end{table}
\end{smaller}

%% file: 08_conclusion.tex
\section{Conclusion}
In this paper, we proposed a method to develop a self-explainable deep learning~(SE-DL) system for a single-class image classification problem. The proposed SE-DL system generates an explanation along with its output class prediction for an image classification problem. Additionally, we leverage the explanation of the proposed deep learning~(DL) system to detect potential class prediction errors of the DL system.
The proposed SE-DL system relies on a selected set of concepts to generate the explanation. A proposed concept selection method is also presented that selects a subset of concepts based on their contribution to the error detection performance of the SE-DL system. 

Furthermore, We evaluated the SE-DL image classification system in the face of adversarial examples. Specifically, we compared the error detection performance of the enhanced self-error detection mechanism using an additional regular classifier with an existing ensemble technique.
We illustrated that given the highest perturbation of adversarial example, the enhanced self-error detection mechanism achieves 97.7\% error detection performance, while the existing ensemble technique achieves 87.5\% error detection performance at a comparable computational complexity and memory size. 